\documentclass[10pt,twocolumn,letterpaper]{article}

\usepackage{wacv}              % To produce the CAMERA-READY version
\definecolor{wacvblue}{rgb}{0.21,0.49,0.74}
\usepackage[pagebackref,breaklinks,colorlinks,allcolors=wacvblue]{hyperref}

\usepackage{graphicx}
\usepackage{booktabs}
\usepackage{bbding}
\usepackage{multirow}
\usepackage{wrapfig}
\usepackage{appendix}
\usepackage{setspace}
\usepackage{float}
\usepackage[table,xcdraw]{xcolor}
\usepackage{makecell}
\usepackage[misc]{ifsym} 
\usepackage{pifont}

\def\wacvPaperID{1751} % *** Enter the WACV Paper ID here
\def\confName{WACV}
\def\confYear{2026}

\title{Structure-Aware Feature Rectification with Region Adjacency Graphs for Training-Free Open-Vocabulary Semantic Segmentation}

\author{
Qiming Huang, Hao Ai, Jianbo Jiao\\
The MIx Group, School of Computer Science\\
University of Birmingham\\
{\tt\small \{qxh366, hxa456\}@student.bham.ac.uk, j.jiao@bham.ac.uk}
}

\begin{document}
\maketitle
\begin{abstract}
Benefiting from the inductive biases learned from large-scale datasets, open-vocabulary semantic segmentation (OVSS) leverages the power of vision-language models, such as CLIP, to achieve remarkable progress without requiring task-specific training. However, due to CLIP’s pretraining nature on image-text pairs, it tends to focus on global semantic alignment, resulting in suboptimal performance when associating fine-grained visual regions with text. This leads to noisy and inconsistent predictions, particularly in local areas. We attribute this to a dispersed bias stemming from its contrastive training paradigm, which is difficult to alleviate using CLIP features alone. To address this, we propose a structure-aware feature rectification approach that incorporates instance-specific priors derived directly from the image. Specifically, we construct a region adjacency graph (RAG) based on low-level features (\eg colour and texture) to capture local structural relationships and use it to refine CLIP features by enhancing local discrimination. Extensive experiments show that our method effectively suppresses segmentation noise, improves region-level consistency, and achieves strong performance on multiple open-vocabulary segmentation benchmarks. Project page: \href{https://qiming-huang.github.io/RAG-OVS/}{https://qiming-huang.github.io/RAG-OVS/}.
\end{abstract}
    
\section{Introduction}

\begin{figure}[h]
    \centering
    \includegraphics[width=\linewidth]{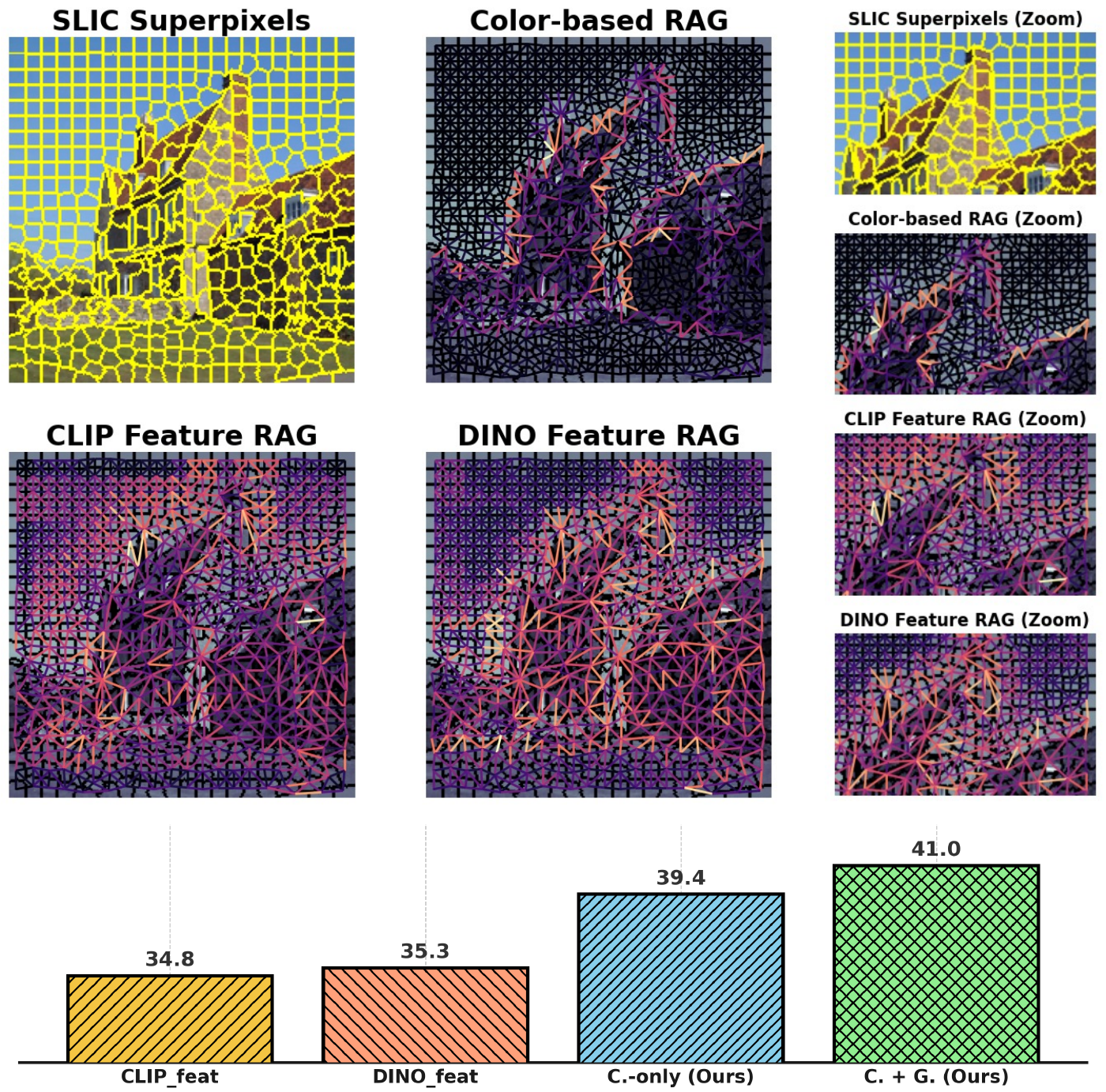} 
    \caption{\textbf{Illustration of the main idea and performance.} High-level feature region adjacency graphs (RAGs) introduce local noise, while low-level colour-based RAGs maintain clean structure. The RAGs built on CLIP~\cite{clip} and DINO~\cite{dino} pretrained features exhibit noisy and inconsistent connectivity in local regions (see zoomed-in areas), when compared to the low-level based one. This highlights the potential of low-level cues for tasks requiring fine-grained local modelling, \eg image segmentation. \textbf{Bottom}: Comparison of average performance across multiple datasets using different features for RAG construction. C.-only: colour-based features, and C. + G.: colour and texture features.}
    \label{fig:example_rag}
\end{figure}

Pretrained vision-language models such as CLIP~\cite{clip} have demonstrated remarkable performance in zero-shot and open-vocabulary recognition tasks. Despite its effectiveness in capturing global image-text alignment, CLIP suffers from notable inductive biases at the local image level, limiting its applicability to fine-grained visual understanding. Specifically, CLIP was trained on image-text pairs without explicit supervision, enforcing implicit alignment between visual regions and textual descriptions. Consequently, it captures coarse semantic correspondences rather than fine-grained regional details, making it less effective for tasks requiring high spatial granularity, such as training-free open-vocabulary semantic segmentation (OVSS). This limitation manifests as noisy and inconsistent region-level predictions in training-free OVSS. As shown in Fig.~\ref{fig:example_rag}, we observe that features extracted from CLIP~\cite{clip} and DINO~\cite{dino} lack clear discrimination across local (superpixel) regions, with high levels of noise and blurred boundaries. In contrast, simple cues such as average colour differences are able to reflect local structural differences more clearly. This motivates us to ask: \textit{Can low-level region-adjacency information be leveraged to guide CLIP toward more localised attention}?

These observations align with the contrastive training paradigm of CLIP, which encourages semantic alignment based on paired image-text data. Since high-resolution images naturally contain more discriminative details, they improve the model’s ability to form stable and structured similarity matrices. However, the inductive bias introduced by CLIP’s global training procedure cannot be easily mitigated using its own representations. As shown in Fig. \ref{fig:example_rag}, CLIP and DINO features are scattered and unstructured in local regions. Fortunately, the image itself inherently provides instance-specific priors that are largely immune to such global alignment biases. Specifically, the Region Adjacency Graph (RAG), constructed purely from low-level cues such as colour and texture, effectively captures spatial relationships between regions without being affected by CLIP’s global feature behaviour. Motivated by this, we propose a structure-aware feature rectification approach that incorporates RAG-based guidance into attention mechanisms. By constructing RAGs from low-level features (\eg colour, GLCM texture statistics), we introduce local structure-aware biases to guide patch-level attention and similarity computation.

Extensive experiments validate that the proposed method enhances training-free open-vocabulary semantic segmentation performance. It improves regional consistency, reduces noise in segmentation outputs, and better preserves fine-grained structures, which are clearly visible in qualitative results (see Fig.~\ref{fig:vis}).

% \begin{figure}[]
%     \centering
%     \includegraphics[width=\linewidth]{src/bounds.pdf} % 预定义占位图片
%     \caption{The analysis of the upper bound on subspace deviation for the Gram matrix of vis-text similarity under perturbation, computed using the Davis–Kahan $\sin \Theta$ theorem. The y-axis represents the normalised upper bound $\frac{\|\Delta\|}{\delta}$, where a lower value indicates greater stability of the sim matrix under perturbations. The x-axis denotes image resolution. As resolution increases, the subspace deviation upper bound decreases, suggesting that higher-res imgs yield more stable vis-text alignment. Different curves correspond to various datasets, showing a consistent trend across multiple eval settings.}
%     \label{fig:sin_}
% \end{figure}

\section{Related Work}
\label{c}
\subsection{Contrastive Language-Image Pre-training}
\label{c1}

% Contrastive Language-Image Pre-training (CLIP)\cite{clip} is a large multi-modal foundation model, which utilizes the contrastive training of aligning visual and text corresponding category features, greatly improves the generalization on unseen samples. Currently CLIP is widely used in Few-Shot/Zero-Shot Learning (FSL/ZSL)\cite{coop,cocoop,maple,rpo,clipes}, Prompt learning\cite{coop,cocoop,maple,rpo} and Out-of-Distribution (OoD)\cite{clipood} tasks. 
% Later, researchers begin to apply CLIP to dense feature tasks\cite{icar,openseed,regionclip,proposalclip} such as semantic segmentation\cite{clipes,cgformer}.

% Li et al.\cite{clipsurgery} elaborate on the inherent noise problem of CLIP and introduce it into the open vocabulary task from the perspective of explainability through self-attention improvement. 
% Unlike pipelines that generally fine-tune pre-trained models on additional data sets, the CLIP encoder often needs to be frozen and cannot be fine-tuned because it needs to maintain alignment with the text feature space\cite{maskclip}. Therefore, researchers currently prefer to use clip directly as an encoder to obtain preliminary features to inherit its excellent generalization ability, and pay more attention to design sophisticated decoders\cite{zegformer,denseclip,catseg,clsclip,mvpseg} to refine the image-level features to adapt to dense feature tasks.

Contrastive Language-Image Pre-training (CLIP)~\cite{clip} is a large-scale multi-modal foundation model that leverages contrastive learning to align visual and textual features, enhancing generalisation on unseen samples. Due to its strong zero-shot capabilities, CLIP has been widely adopted in Few-Shot/Zero-Shot Learning (FSL/ZSL)~\cite{coop,cocoop,maple,rpo,clipes}, Prompt Learning~\cite{coop,cocoop,maple,rpo}, and Out-of-Distribution (OoD) detection tasks~\cite{clipood}. 

More recently, researchers have extended CLIP to dense prediction tasks~\cite{icar,openseed,regionclip,proposalclip}, such as semantic segmentation~\cite{clipes,cgformer}. However, a major challenge in utilising CLIP is the inherent noise in its features. Li \etal~\cite{clipsurgery} analyse this issue from an explainability perspective and propose self-attention improvements to enhance CLIP’s performance in open-vocabulary tasks. 

Unlike conventional pipelines that fine-tune pre-trained models on additional datasets, CLIP’s encoder is typically kept frozen to maintain its alignment with the text feature space~\cite{maskclip}. As a result, researchers tend to use CLIP directly as an encoder to extract preliminary features while focusing on designing sophisticated decoders~\cite{zegformer,denseclip,catseg,clsclip,mvpseg} to refine image-level representations for dense prediction tasks.

\subsection{Open-Vocabulary Semantic Segmentation}
Open-vocabulary semantic segmentation (OVSS) extends segmentation ~\cite{tian2020prior, peng2023hierarchical, tian2022generalized} and refers to segmenting semantic regions via textual names or descriptions for the open world without any mask annotations.
Early works~\cite{maskclip} verify the importance of modal alignment in CLIP, and common downstream fine-tuning may destroy its generalisation ability. 
MaskCLIP~\cite{maskclip} attempts to improve the Vision Transformer (ViT)~\cite{vit} structure of CLIP to allow the model to obtain coarse feature localisation, and combines transductive learning to improve performance.
CLIP-Surgery~\cite{li2025closer} analyses the difficulty of the current semantic segmentation task introduced by CLIP from the perspective of image-text noise, and makes certain improvements to the model using the idea of self-attention.
SCLIP~\cite{sclip} inherits the idea of self-attention from MaskCLIP and directly adapts the improved CLIP structure to the semantic segmentation task.

Both CLIP-Surgery and SCLIP utilise the idea of self-attention to improve CLIP, while only CLIP-Surgery mentions the noise problem caused by the open category of text.
None of them explores and analyses why CLIP lacks the semantic correlation between patches.
Our work complements this point that it is the global patch formed during the attention interaction between [CLS] token and patches that leads to this.

Beyond architectural improvements, recent work scrutinizes OVS evaluation protocols regarding task ambiguity. Huang et al.~\cite{huang2025revisit} argue that rigid pixel-wise metrics contradict the open-world premise by penalizing plausible synonyms (\eg, `sofa' vs. `couch'). They propose a mask-wise evaluation protocol, demonstrating that mitigating category ambiguity significantly enhances model capabilities and suggesting a need for evolved benchmarks.

Regarding methodology, recent approaches leverage CLIP as an encoder within a "mask generation and classification" pipeline, inspired by MaskFormer~\cite{maskformer} and Mask2Former~\cite{mask2former}. These methods utilise pixel and query decoders to refine features and generate masks via query embeddings. By calculating the similarity between these embeddings and text prompts, the model weights query masks to produce final object boundaries and categories.

% One of the fundamental challenges in this task is maintaining the alignment between vision and language modalities while ensuring generalisation beyond the training distribution. Early explorations~\cite{maskclip} emphasise the significance of preserving CLIP’s multi-modal alignment, as naive fine-tuning often degrades its zero-shot capabilities. To address this, MaskCLIP~\cite{maskclip} modifies the Vision Transformer (ViT)~\cite{vit} structure within CLIP to improve spatial feature localisation. By integrating transductive learning, it enhances segmentation performance by refining the representation of visual concepts.

% SCLIP~\cite{sclip} builds upon MaskCLIP by further improving the self-attention mechanism to better capture semantic information across image regions. On the other hand, CLIP-Surgery introduces an alternative perspective by analyzing the noise introduced by open-category text inputs. It mitigates this issue through targeted modifications in self-attention while also addressing the inefficiencies in CLIP’s ability to align image-text embeddings for segmentation.

% Despite their advancements, none of these approaches have thoroughly investigated the lack of semantic coherence among patches in CLIP-based models. Our work complements this gap by identifying the role of global patch interactions—particularly the attention mechanism between the [CLS] token and image patches—as a key factor influencing semantic consistency.

\subsection{Training-free OVSS} Trident~\cite{shi2024harnessing} proposes a training-free framework that addresses CLIP’s resolution limitation in semantic segmentation through a splice-then-segment approach. Trident first splices features extracted by CLIP and DINO from sub-images, then leverages the Segment Anything Model (SAM) for global aggregation, expanding the receptive field and improving segmentation performance. The kNN-CLIP~\cite{gui2024knn} proposes a training-free approach for open-vocabulary continual segmentation that mitigates catastrophic forgetting. Instead of traditional continual training, kNN-CLIP augments the model with a database of instance embeddings, enabling segmentation methods to adapt to growing vocabularies without retraining or high memory costs.  These methods primarily modify the internal attention structure of CLIP-like models to better capture relationships between image regions and textual descriptions.

In contrast, our approach takes a different direction by directly modifying the visual patch embeddings instead of adjusting attention maps. Specifically, we improve the accuracy of the visual patch-text embedding similarity matrix, ensuring a more precise alignment between visual and textual representations. By refining the embedding space at the patch level, our method enhances feature interaction and boosts segmentation performance, complementing and surpassing attention-based optimisation strategies.

% \section{Method}

% \begin{figure*}[h]
%     \centering
%     \includegraphics[width=\linewidth]{src/model.pdf} % 预定义占位图片
%     \caption{Illustration of the proposed feature rectification. The model takes an input image and partitions it into patches, each of which is transformed into patch feature embeddings. These embeddings are refined through the integration of text feature embeddings, represented as $[v_1, v_2, ..., v_n]$, to enhance semantic understanding. A region adjacency graph is constructed based on neighborhood relationships between patches, where the variance of edge weights guides the correction of ambiguous feature representations. The core component of the model, Spherical Linear Interpolation (Slerp)-based Feature Rectification, enables smooth and structure-preserving rectification between patch embeddings. This operation interpolates between the original patch feature embedding $p_i$ and the refined embedding $v_i$ along a spherical manifold, ensuring that the final representation remains semantically aligned while maintaining spatial coherence. The rectified features are then used to generate the final segmentation prediction, which exhibits improved accuracy in capturing object boundaries and contextual dependencies.}
%     \label{fig:model}
% \end{figure*}

% \subsection{Limitation of CLIP in Segmentation}

% xxx

\section{Preliminaries of Training-free OVSS}

Training-free OVSS aims to segment an image into meaningful regions by assigning semantic labels given arbitrary vocabulary, without requiring extra training. Instead of learning a segmentation model with annotated data, this approach leverages large pretrained vision-language models, such as CLIP, to directly match visual features with text embeddings through similarity computations. The visual patches embedding $\{\mathbf{v}_i\}_{i=1}^{N}$, where each patch $\mathbf{v}_i$ is represented by a feature embedding of dimension $1 \times \mathbb{R}^{D}$, extracted from the vision encoder of CLIP. The text embedding $\{\mathbf{t}_j\}_{j=1}^{M}$ is obtained from the text encoder, where each $\mathbf{t}_j$ corresponds to a text and is also represented as a feature embedding of dimension $1 \times \mathbb{R}^{D}$. The core idea is to compute the cosine similarity between each visual patch feature and all text embeddings:
\begin{equation}\label{sim}
s_{i,j} = \frac{\langle \mathbf{v}_i, \mathbf{t}_j \rangle}{\| \mathbf{v}_i \| \| \mathbf{t}_j \|},
\end{equation}
where $ s_{i,j} $ is the similarity score between visual patch $v_i$ and text embedding $v_j$. The semantic label for each visual patch is assigned based on the highest similarity score:
\begin{equation}\label{prediction}
\hat{y}_i = \arg\max_{j} s_{i,j},
\end{equation}
where $ \hat{y}_i $ denotes the predicted semantic label for patch $\mathbf{v}_i$.

% However, due to the region distracting biases in CLIP. The predicted segmentations often exhibit issues of noise and regional inconsistencies. To address this issue, we propose a feature rectification strategy that interpolates the original feature embedding to a refined feature embedding for each patch $\mathbf{v}_i$, aiming to mitigate the impact of CLIP’s inductive bias.

\section{Structure-Aware Feature Rectification}

Due to CLIP's global training paradigm on image-text pairs, it lacks the capability for fine-grained local alignment~\cite{naclip, bai2024self, cliptrace, lan2024proxyclip}, resulting in structural inconsistency and noisy predictions when directly applied to segmentation. This issue is especially pronounced in training-free open-vocabulary semantic segmentation, where no additional data is available for model adaptation. To mitigate this, our method leverages a region adjacency graph (RAG) constructed from image low-level features to enhance structural awareness. It comprises two key modules: RAG-guided Attention, which introduces a structure-aware bias into CLIP’s attention mechanism to encourage local semantic consistency; and Similarity Fusion, which refines cross-modal similarity computation to suppress noisy matches.

\subsection{RAG-guided Attention} 
% A Region Adjacency Graph (RAG) is a graph-based representation that captures the spatial relationships between image regions. Each node in the graph corresponds to a low-level region (e.g., a superpixel), and edges represent adjacency between regions. By encoding both the appearance and structural proximity of regions, the RAG provides a compact yet informative structure that reflects the local layout of the image. 

% However, traditional Region Adjacency Graphs (RAGs) are typically constructed based on the average color differences between superpixel regions. While this provides clear local structural cues—as illustrated in Fig. \ref{fig:example_rag}—color differences alone are insufficient for robust region discrimination. In real-world scenarios, color ambiguity often arises, such as between a ``
% white toilet'' and a ``white wall''. To build a more robust RAG, we incorporate not only color differences but also texture information. Specifically, we compute the Gray-Level Co-occurrence Matrix (GLCM) for each region and extract several statistical features, including contrast, homogeneity, energy, and correlation, to better characterize the appearance of each region.

Region adjacency graph (RAG) is a graph-based representation that captures the spatial relationships between image regions. Formally, a RAG is defined as an undirected graph $G = (V, E)$, where each node $v_i \in V$ corresponds to a low-level region $R_i$ (\eg a superpixel), and an edge $e_{ij} \in E$ exists if regions $R_i$ and $R_j$ are spatially adjacent in the image. By encoding both the appearance and structural proximity of regions, the RAG provides a compact yet informative structure that reflects the local layout of the image.

\paragraph{RAG construction.} However, traditional RAGs are typically constructed based on the average colour differences between adjacent superpixel regions. The weight of each edge $e_{ij}$ is defined as:

\begin{equation}
    w_{ij}^{\text{color}} = \| \mu_i - \mu_j \|_2,
\end{equation}
where $\mu_i$ and $\mu_j$ denote the mean RGB colour vectors of regions $R_i$ and $R_j$, respectively. While this formulation provides clear local structural cues---as illustrated in Fig.~\ref{fig:example_rag}--- colour differences alone are insufficient for robust region discrimination. In real-world scenarios, colour ambiguity often arises, such as between a ``white toilet'' and a ``white wall''. To build a more robust RAG, we incorporate not only colour differences but also texture information. Specifically, for each region $R_i$, we compute the \textit{Grey-Level Co-occurrence Matrix (GLCM)} $P_i$ and extract several statistical features from it. The edge weight is redefined as a combination of colour and texture similarities:

\begin{equation}
    w_{ij} = w_{ij}^{\text{color}} + w_{ij}^{\text{texture}},
\end{equation}
where $w_{ij}^{\text{texture}}$ is computed using the GLCM-based feature difference:
\begin{equation}\label{dis}
    w_{ij}^{\text{texture}} = \sum_{k} \left| f_i^{(k)} - f_j^{(k)} \right|,
\end{equation}
in which $f_i^{(k)}$ represents the $k$-th texture feature (\eg contrast, homogeneity, energy, correlation) extracted from region $R_i$'s GLCM. These features are defined as follows:  
contrast: $ \text{Contrast} = \sum_{m,n} (m - n)^2 P_i(m,n) $,  
homogeneity: $ \text{Homogeneity} = \sum_{m,n} \frac{P_i(m,n)}{1 + |m - n|} $,  
energy: $ \text{Energy} = \sum_{m,n} P_i(m,n)^2 $,  
and correlation: $ \text{Correlation} = \frac{\sum_{m,n} (m - \mu_m)(n - \mu_n) P_i(m,n)}{\sigma_m \sigma_n} $, where $P_i(m,n)$ denotes the normalized co-occurrence probability at position $(m,n)$, and $\mu_m, \mu_n, \sigma_m, \sigma_n$ are the means and standard deviations of the marginal distributions of $P_i$.

\begin{figure}
    \centering
    \includegraphics[width=\linewidth]{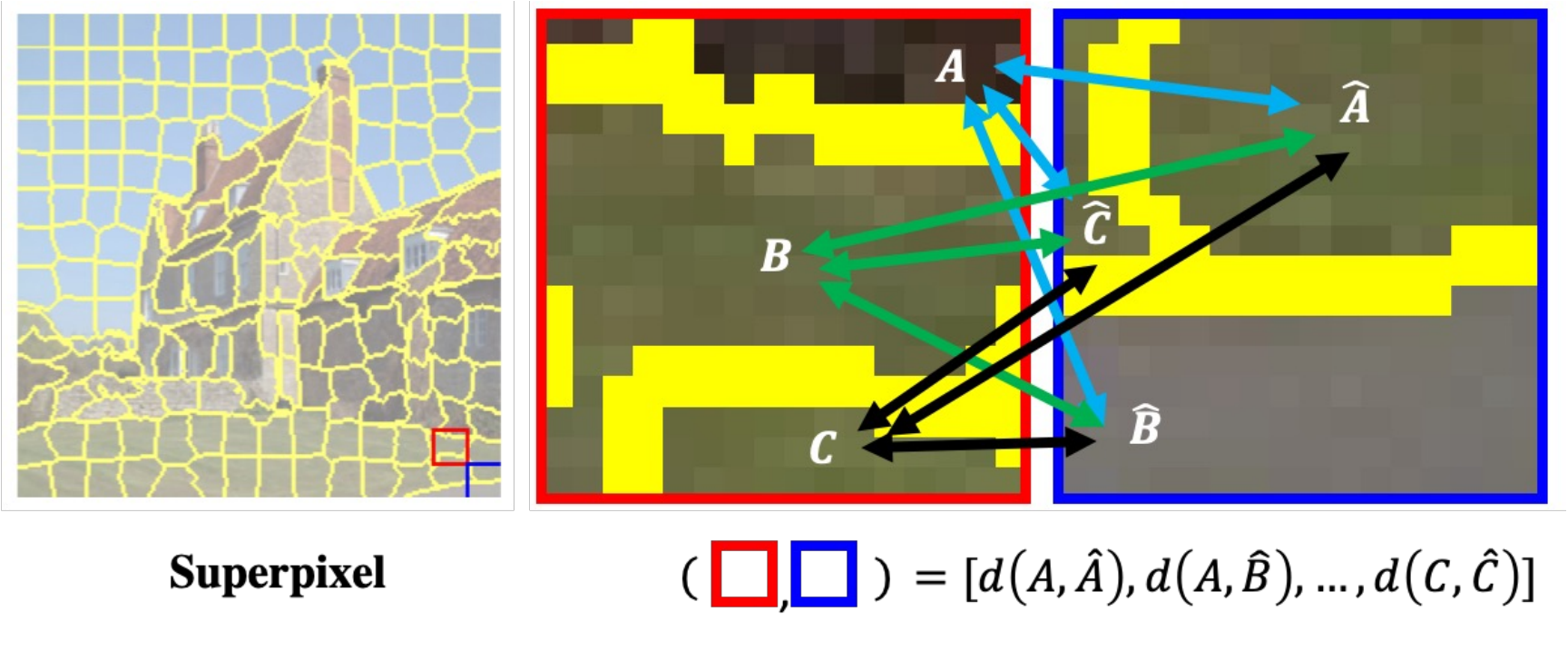}
    \caption{\textbf{Illustration of superpixel-to-patch encoding.} The distance between two patches is first represented as a list of all pairwise superpixel regions $(\textcolor{red}{\boldsymbol{\Box}},\textcolor{blue}{\boldsymbol{\Box}})$, then the patch distance is computed from this list using Eq. \ref{dis_patch}.}
    \label{fig:superpixel2patch}
\end{figure}

\paragraph{Superpixel-aligned patch encoding.}
Another challenge lies in the mismatch between the superpixel-based RAG and the patch-based tokenisation used in transformers, where inputs are typically divided into fixed-size square patches. To address this, we design a mechanism that preserves the structural advantages of superpixels—such as their ability to align flexibly with object boundaries—while enabling compatibility with patch-wise representations required by standard transformer attention. As illustrated in Fig.~\ref{fig:superpixel2patch}, for two adjacent patches, denoted as $\textcolor{red}{\boldsymbol{\Box}}$ and $\textcolor{blue}{\boldsymbol{\Box}}$, the computation of their edge weight is based on all pairwise distances between the superpixel regions contained within each patch. Specifically, let patch $i$ contain superpixels $\{s_1^i, s_2^i, \dots, s_m^i\}$ and patch $j$ contain superpixels $\{s_1^j, s_2^j, \dots, s_n^j\}$. We compute the pairwise distances:

\begin{equation}
    \mathcal{D}_{ij} = \left\{ d(s_p^i, s_q^j) \mid s_p^i \in i, \; s_q^j \in j \right\},
\end{equation}
where $d(\cdot, \cdot)$ is the distance function defined in Eq.~\ref{dis}. Therefore the computed edge weight $w_{ij}$ is a list rather than a scalar, \ie

\begin{equation}
    w_{ij} = \left[ d(s_1^i, s_1^j), d(s_1^i, s_2^j), \dots, d(s_m^i, s_n^j) \right].
\end{equation}

To preserve the structural variations within each patch, we compute the mean and variance of $\mathcal{D}_{ij}$, and use them as the final edge weight representation between patch $i$ and patch $j$:

\begin{equation}\label{dis_patch}
    \mu_{ij} = \frac{1}{|\mathcal{D}_{ij}|} \sum_{d \in \mathcal{D}_{ij}} d, \quad
\sigma^2_{ij} = \frac{1}{|\mathcal{D}_{ij}|} \sum_{d \in \mathcal{D}_{ij}} (d - \mu_{ij})^2.
\end{equation}

We then define the final edge weight list as (we use the standard deviation $\sigma_{i,j}$):

\begin{equation}
    w_{ij}^{\text{final}} = [\mu_{ij}, \sigma_{ij}].
\end{equation}

\begin{figure}
    \centering
    \includegraphics[width=\linewidth]{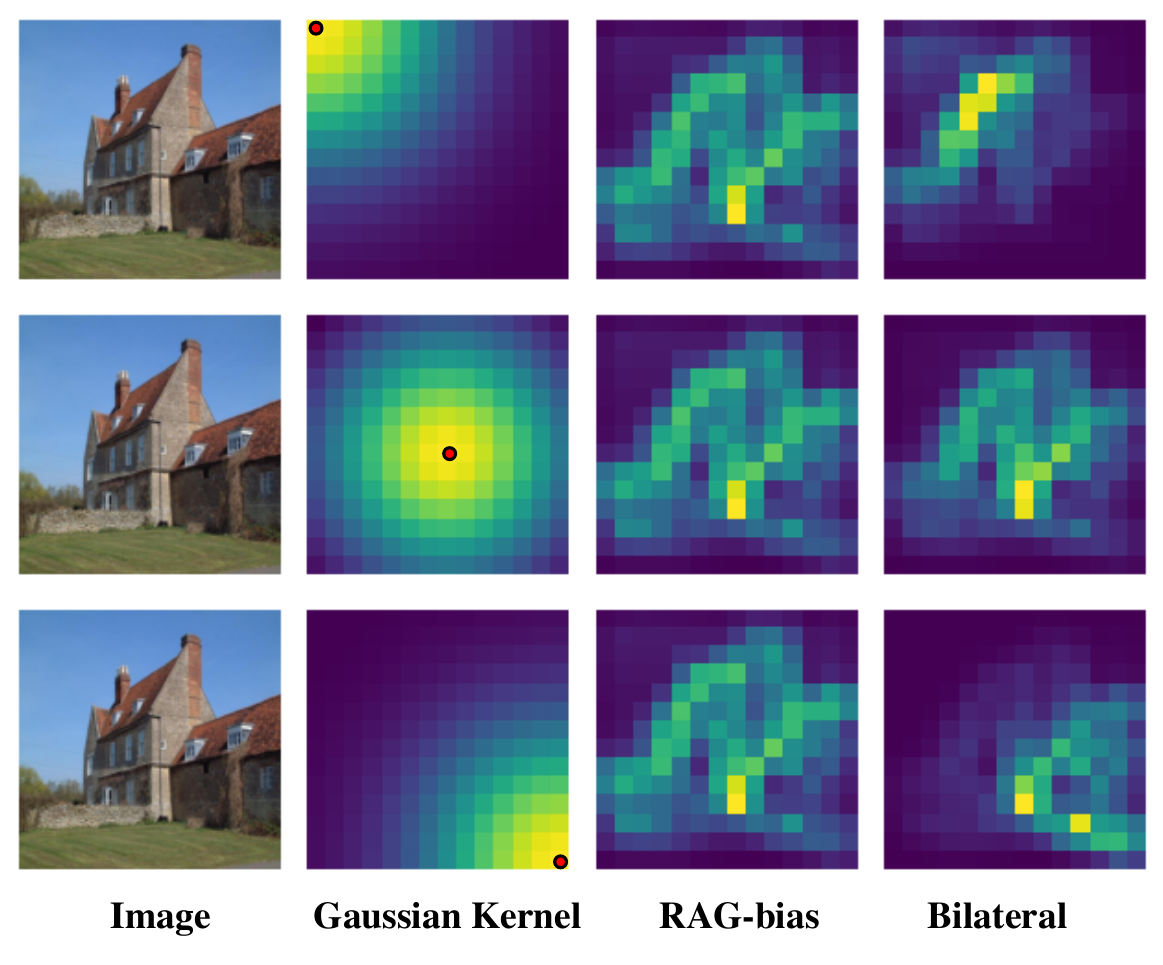}
    \caption{\textbf{Illustration of different attention bias mechanisms.} The first column shows the input images. The second column visualises the traditional Gaussian kernel, which models spatial proximity in a local window. The third column shows the RAG-bias computed from the Region Adjacency Graph (RAG), capturing structural relationships between neighbouring regions. The fourth column combines both the Gaussian kernel and the RAG-bias to form a bilateral attention bias, which accounts for both spatial distance and local structure.}
    \label{fig:rag-bias}
\end{figure}

\paragraph{RAG-guided attention via RAG bias.} We leverage the constructed Region Adjacency Graph (RAG) to compute a structure-aware prior, referred to as the \textit{RAG bias}, which serves as a local structural constraint in the attention mechanism. As illustrated in Fig.~\ref{fig:rag-bias} (third column), the RAG bias is calculated for each token (patch) based on the topology of its local neighbourhood in the RAG.

Specifically, for a node (patch) $i$, we consider its adjacent neighbours $\mathcal{N}(i)$—defined either as 4-connected (cross-shaped) or 8-connected neighbours. For each neighbour $j \in \mathcal{N}(i)$, we use the final edge weight $w_{ij}^{\text{final}} = [\mu_{ij}, \sigma_{ij}^2]$. The RAG bias $b_{ij}$ is then computed by averaging the structural affinities from node $i$'s neighbourhood:

\begin{equation} \label{neh}
    b_{ij} = \frac{1}{|\mathcal{N}(i)|} \sum_{k \in \mathcal{N}(i)} \left( \mu_{ik} + \sigma_{ik} \right).
\end{equation}

Although the RAG bias $b_{i,j}$ effectively encodes structural context by aggregating information from a node's local neighbourhood, it is fixed across all positions within the same image. This static nature makes it insufficient for capturing the pairwise relationships required in self-attention, where different attention weights are computed between every pair of tokens. To address this limitation, we draw inspiration from bilateral filtering and introduce a more flexible bias mechanism, Bilateral Bias, that combines both spatial proximity and structural similarity. Specifically, we compute a spatial Gaussian kernel $g(i,j)$ between any two positions $i$ and $j$:

\begin{figure}
    \centering
    \includegraphics[width=\linewidth]{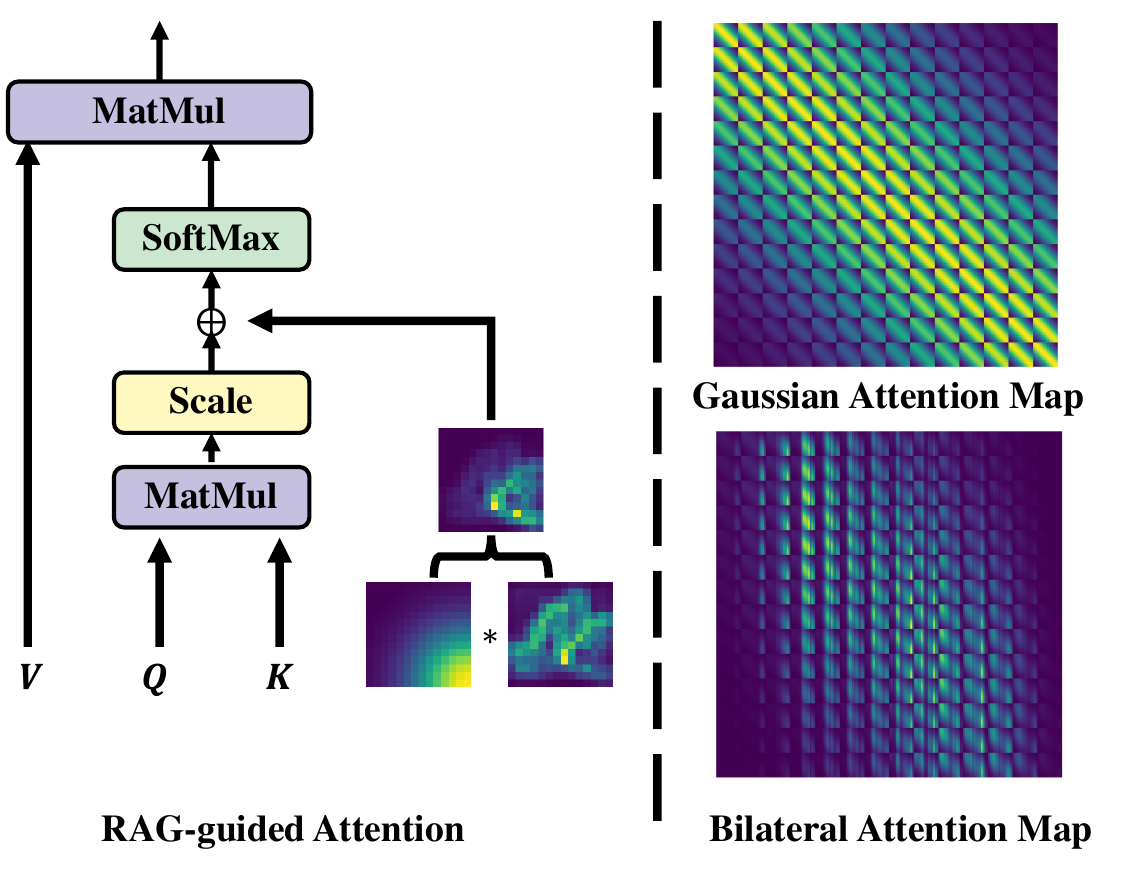}
    \caption{\textbf{Overview of the proposed RAG-guided attention mechanism.} The bilateral attention bias is computed by combining a spatial Gaussian kernel with a structure-aware RAG-bias. This combined bias is integrated into the attention weights to enhance structural sensitivity. \textbf{Right}: visualisation of the Gaussian and bilateral attention maps.}
    \label{fig:rag-bias-atten}
\end{figure}

\begin{equation}
    g(i,j) = \exp\left( -\frac{\| p_i - p_j \|^2}{2\sigma^2} \right),
\end{equation}
where $p_i$ and $p_j$ denote the 2D coordinates of patches $i$ and $j$, and $\sigma$ controls the spatial range.

We then define the bilateral bias $B_{ij}$ as the product of the spatial kernel and the structural RAG bias:

\begin{equation}
    B_{ij} = g(i,j) \cdot \exp\left( b_{i,j}\right)
\end{equation}
as shown in Fig. \ref{fig:rag-bias} (fourth column). Then the final biased attention is computed as:
\begin{equation}
    atten_{ij}^{\text{biased}} = \text{softmax}_j(\frac{\mathbf{q}_i \cdot \mathbf{k}_j}{\sqrt{d}} + B_{ij}).
\end{equation}

% As shown in Fig.~\ref{fig:rag-bias}, the second column shows the spatial Gaussian kernel $g(i,j)$, while the fourth column shows the resulting bilateral bias $B_{ij}$ that captures both local spatial proximity and structure-aware semantic similarity.

The overall of this RAG-guided attention process is shown in Fig. \ref{fig:rag-bias-atten}.

% In the self-attention mechanism, the input sequence of token embeddings is linearly projected using the transformation matrix $W^{qkv}$ to obtain three distinct $d$-dimensional vectors: query ($\mathbf{q}$), key ($\mathbf{k}$), and value ($\mathbf{v}$). To compute the attention, the similarity between all token pairs is measured. Specifically, for a patch located at position $(i,j)$, the similarity between $\mathbf{q}_{ij}$ and all $\mathbf{k}_{mn}$ vectors (where $m \in \{1,2,\dots,h\}$ and $n \in \{1,2,\dots,w\}$) is calculated using the dot product. This similarity is scaled by $1/\sqrt{d}$ and passed through a softmax to obtain a distribution used to weight the value vectors. The resulting weighted sum provides the attention output, which is then linearly transformed using another matrix $W^{\circ}$. The full computation for the patch at $(i,j)$ can be formalized as:

% \begin{align}
% [\mathbf{q}, \mathbf{k}, \mathbf{v}] &= \mathbf{Z} W^{qkv}, \\
% \text{sim}_{ij} &= \frac{\mathbf{k} \mathbf{q}_{ij}}{\sqrt{d}}, \\
% A_{ij} &= \text{softmax}(\text{sim}_{ij}) \mathbf{v}, \\
% \text{SA}(\mathbf{Z})_{ij} &= A_{ij} W^{\circ}.
% \end{align}

% From here onward, we refer to $\text{softmax}(\text{sim}_{ij})$ as the \textit{attention map} of position $(i,j)$, and $\text{sim}_{ij}$ as its \textit{logits}. Although multi-head attention is generally employed, this explanation considers only the single-head case for clarity.

\subsection{Similarity Fusion Module} 

Although introducing a bias that emphasises local structures helps the model attend to fine-grained region boundaries and maintain local consistency, it may also cause the model to respond to irrelevant local noise, such as background textures, illumination variations, or boundary artefacts. To mitigate this, we propose similarity fusion.

Specifically, given the original visual-textual similarity matrix $S_{i,j}$ defined in Eq.~\ref{sim}, we compute a refined similarity matrix $\tilde{S}_{i,j}$ by first applying a Gaussian kernel to smooth the visual features. Let $\hat{\mathbf{v}}_i$ denote the smoothed visual feature at position $i$; we can then compute the cosine similarity between the smoothed visual features and the text feature $\mathbf{t}$:

\begin{equation}
    \tilde{S}_{i,j} = \frac{\hat{\mathbf{v}}_i \cdot \mathbf{t}_j}{\|\hat{\mathbf{v}}_i\| \, \|\mathbf{t}_j\|}.
\end{equation}

Finally, we fuse the original and smoothed similarities using geometric mean fusion:

\begin{equation}
    S^{\text{fused}}_{i,j} = \left( \tilde{S}_{i,j} \right)^\alpha \cdot \left( S_{i,j} \right)^{1 - \alpha}.
\end{equation}

\begin{table*}[] \centering
\caption{\textbf{Quantitative results on various OVSS benchmarks.} Our method consistently improves different CLIP-based baselines across all datasets, showing its generality and effectiveness. Best performance in \textbf{bold}.}
\label{tab:main}
\begin{tabular}{l|c|cccccccc!{\vrule width 2pt}c}
\toprule
\textbf{Method}               & \textbf{Venue}                & \textbf{V21}  & \textbf{PC60} & \textbf{C-Obj} & \textbf{V20}  & \textbf{PC59} & \textbf{Stuff} & \textbf{City} & \textbf{ADE}  & \textbf{Avg}  \\ \midrule
CLIP~\cite{clip}              & {\color[HTML]{C0C0C0} ICML’21}      & 16.4          & 8.4           & 5.6            & 41.9          & 9.2           & 4.4            & 5.0           & 2.9           & 11.7          \\
GroupViT~\cite{groupvit}      & {\color[HTML]{C0C0C0} CVPR’22}      & 52.3          & 18.7          & 27.5           & 79.7          & 18.5          & 23.4           & 10.4          & 15.3          & 30.7          \\
MaskCLIP~\cite{maskclip}      & {\color[HTML]{C0C0C0} ECCV’22}      & 43.4          & 23.2          & 20.6           & 74.9          & 26.4          & 16.7           & 24.9          & 11.9          & 30.3          \\
Reco~\cite{reco}              & {\color[HTML]{C0C0C0} NeurIPS’22}   & 25.1          & 19.9          & 15.7           & 57.7          & 21.6          & 22.3           & 11.2          & 14.8          & 23.5          \\
OVDiff~\cite{ovdiff}          & {\color[HTML]{C0C0C0} ECCV’24}      & 66.3          & 29.7          & 34.6           & 80.9          & 32.9          & 20.3           & 23.4          & 14.1          & 37.8          \\ \hline
CLIP-Surgery~\cite{clipsurgery} & {\color[HTML]{C0C0C0} Pattern Recognition} & 59.0 & 30.1 & 30.2 & 80.1 & 33.9 & 22.1 & 31.8 & 15.8 & 37.9 \\
\rowcolor[HTML]{ECF4FF} 
\hspace{1em}\textbf{$\hookrightarrow$ + Ours} & {\color[HTML]{C0C0C0} -} & \textbf{61.1} & \textbf{32.2} & \textbf{31.8}  & \textbf{81.3} & \textbf{34.8} & \textbf{23.6}  & \textbf{33.5} & \textbf{17.4} & \textbf{39.4} \\
SCLIP~\cite{sclip}            & {\color[HTML]{C0C0C0} ECCV’24}      & 59.1          & 30.4          & 30.5           & 80.4          & 34.2          & 22.4           & 32.2          & 16.1          & 38.2          \\
\rowcolor[HTML]{ECF4FF} 
\hspace{1em}\textbf{$\hookrightarrow$ + Ours} & {\color[HTML]{C0C0C0} -}                   & \textbf{61.9} & \textbf{32.9} & \textbf{32.3}  & \textbf{81.8} & \textbf{35.0} & \textbf{24.1}  & \textbf{33.9} & \textbf{18.1} & \textbf{39.9} \\
CLIPtrace~\cite{cliptrace}    & {\color[HTML]{C0C0C0} ECCV’24}      & 53.0          & 30.8          & 33.8           & 81.2          & 35.0          & 24.1           & 35.0          & 17.0          & 38.7          \\
\rowcolor[HTML]{ECF4FF} 
\hspace{1em}\textbf{$\hookrightarrow$ + Ours} & {\color[HTML]{C0C0C0} -}                   & \textbf{56.2} & \textbf{32.1} & \textbf{35.2}  & \textbf{83.8} & \textbf{36.4} & \textbf{25.1}  & \textbf{36.6} & \textbf{18.5} & \textbf{40.5} \\
NACLIP~\cite{naclip}          & {\color[HTML]{C0C0C0} WACV’25}      & 58.9          & 32.2          & 33.2           & 79.7          & 35.2          & 23.3           & 35.5          & 17.4          & 39.4          \\
\rowcolor[HTML]{ECF4FF} 
\hspace{1em}\textbf{$\hookrightarrow$ + Ours} & {\color[HTML]{C0C0C0} -}                   & \textbf{60.3} & \textbf{33.5} & \textbf{34.6}  & \textbf{81.2} & \textbf{36.0} & \textbf{25.7}  & \textbf{36.8} & \textbf{19.1} & \textbf{40.9} \\
ProxyCLIP~\cite{lan2024proxyclip} & {\color[HTML]{C0C0C0} ECCV’24}      & 61.3          & 35.3          & 37.5           & 80.3          & 39.1          & 26.5           & 38.1          & 20.2          & 42.3          \\
\rowcolor[HTML]{ECF4FF} 
\hspace{1em}\textbf{$\hookrightarrow$ + Ours} & {\color[HTML]{C0C0C0} -}                   & \textbf{62.9} & \textbf{36.6} & \textbf{38.9}  & \textbf{82.1} & \textbf{39.8} & \textbf{27.7}  & \textbf{40.1} & \textbf{21.1} & \textbf{43.6} \\ \bottomrule
\end{tabular}
\end{table*}

\section{Experiment}

\subsection{Implementation Details} 

\paragraph{Datasets.}
We evaluate our method on the following segmentation benchmarks, whose names are abbreviated (in parentheses) to conserve table space:
PASCAL VOC 2012 (V21) \cite{voc12},
ADE20K-150 (ADE) \cite{ade},
PASCAL Context (PC60) \cite{pascalcontext},
COCO-Stuff (C-Stf) \cite{coco},
Cityscapes (City) \cite{cityscapes},
COCO-Object (C-Obj) \cite{actualcoco}.
Additionally, alongside the original benchmarks on these datasets, we follow \cite{sclip} and evaluate on variants of PASCAL VOC 2012 (V20) and PASCAL Context (PC59) in which the background class is removed from the evaluation.

% Furthermore, input images are resized to have a shorter side of 336 (560 for Cityscapes \cite{cityscapes}, because of high-resolution images), and following prior works \cite{sclip,barsellotti2024fossil,clipsurgery,tcl,groupvit}, a slide inference is carried out with a $224 \times 224$ window and a stride of 112.

\paragraph{Baselines.} 
We compare our method to a set of relevant works in OVSS, including: MaskCLIP~\cite{maskclip}, ReCo~\cite{reco}, GroupVit~\cite{groupvit}, SCLIP~\cite{sclip}, OVDiff~\cite{ovdiff}, CLIPtrace~\cite{cliptrace}, NACLIP~\cite{naclip}, and ProxyCLIP~\cite{lan2024proxyclip}. It is worth noting that, for the sake of fair comparison, none of the methods, including our baselines, involve any post-processing during evaluation. This includes commonly used techniques such as Conditional Random Fields (CRF), multi-scale testing, mask refinement, or other enhancement strategies. All methods are evaluated based on their raw model outputs to ensure a fair and consistent comparison. Specifically, we adopt SCLIP, CLIPtrace, NACLIP, and ProxyCLIP as our baselines and integrate our proposed module into their original frameworks. All other settings strictly follow those described in the respective original papers.

\paragraph{Implementation details.}
All experiments are conducted on a single NVIDIA RTX 4090 GPU. We adopt mean Intersection-over-Union (mIoU) as the evaluation metric across all experiments. For the Similarity Fusion Module, we set the weighting parameter $\alpha$ to 0.6, selected based on the best performance on cocostuff171-val. The Gaussian kernel used in the module has a kernel size of 3 and a standard deviation $\sigma$ of 3. SLIC~\cite{achanta2012slic} is used as our default superpixel method, with n\_segments=300 and compactness=10. More details about the hypaparameter sensitivity analysis can be found in Supplementary Material Section~\ref{s1}.

\subsection{Results}

% In this section, we first present a comparison of our method with baseline models on multiple open-vocabulary segmentation benchmarks, as shown in Table \ref{res_main}. Our feature rectification (Feat. Rect.) consistently improves the performance of CLIP-based segmentation models across all datasets. Specifically, applying Feat. Rect. to CLIP results in a significant improvement, increasing the average mIoU from 11.04\% to 26.22\%, demonstrating the effectiveness of our approach in refining feature alignment. A similar trend is observed for CLIP-Surgery, where our method enhances the PC59 dataset performance from 29.3\% to 31.08\%, and for SCLIP, which improves from 31.50\% to 32.57\% after feature rectification. Moreover, the performance gains are consistently observed across different datasets, highlighting the robustness and generalizability of our method. Notably, CLIPtrase, a state-of-the-art training-free segmentation model, benefits from our approach, improving from 33.53\% to 34.50\% on average, indicating that our rectification method can enhance existing models without requiring additional training. These results confirm that leveraging feature rectification can effectively mitigate the limitations of CLIP-based models, leading to improved segmentation accuracy and better alignment between visual features and textual categories in open-vocabulary segmentation tasks.

Table~\ref{tab:main} presents quantitative results on various OVSS benchmarks, comparing our method with several state-of-the-art baselines. As shown, integrating our proposed module into different CLIP-based models consistently improves performance across all datasets. Notably, our approach yields gains on challenging datasets such as ADE20K, Cityscapes, and PC60, regardless of the baseline model. These improvements validate the effectiveness and generality of our method in enhancing open-vocabulary semantic segmentation performance. Specifically, our method boosts the average mIoU by +1.8 on SCLIP (38.2 → 40.0), +1.8 on CLIPtrace (38.7 → 40.5), +1.5 on NACLIP (39.4 → 40.9), and +1.4 on ProxyCLIP (42.3 → 43.7). These results highlight the generality and effectiveness of our approach in enhancing training-free CLIP-based open-vocabulary segmentation models.

\subsection{Ablation Study}

In this section, we conduct extensive ablation studies to analyse our model. Unless otherwise specified, the baseline used in these ablation studies is NACLIP~\cite{naclip}.

\begin{table}[]\centering
\caption{Component ablation results based on the NACLIP model.}
\label{tab:ablation_com}
\resizebox{\linewidth}{!}{
\begin{tabular}{cc|cccccccc}
\toprule
\multicolumn{2}{c|}{\textbf{Method}}             & \textbf{V21}  & \textbf{PC60} & \textbf{C-Obj} & \textbf{V20}  & \textbf{PC59} & \textbf{Stuff} & \textbf{City} & \textbf{ADE}           \\ \midrule
\rowcolor[HTML]{EFEFEF} 
\multicolumn{2}{c|}{\cellcolor[HTML]{EFEFEF}w/o} & 58.9          & 32.2          & 33.2          & 79.7          & 35.2          & 23.3          & 35.5          & 17.4          \\ \hline
\cellcolor[HTML]{FFCE93}SimFusion               & \cellcolor[HTML]{ECF4FF}RAG-bias               &               &               &               &               &               &               &               &               \\ \midrule
\ding{52}               & \ding{55}              & 59.4 & 32.6 & 33.5 & 80.5 & 35.9 & 24.2 & 36.0 & 18.1 \\
\ding{55}               & \ding{52}              & 60.0          & 33.1          & 34.2          & 81.0          & 36.1          & 25.5          & 36.7          & 19.0          \\
\ding{52}               & \ding{52}              & \textbf{60.2} & \textbf{33.4} & \textbf{34.4} & \textbf{81.0} & \textbf{36.2} & \textbf{25.8} & \textbf{36.9} & \textbf{19.2} \\ \bottomrule
\end{tabular}}
\end{table}

\begin{table}[]
\caption{\textbf{Performance comparison using different feature types to construct RAG edges.} The top half shows results under standard input conditions, while the bottom half (marked with \ding{96}) represents experiments with colour perturbations to evaluate robustness. \textit{C.-only} denotes colour-only input, and \textit{C. + G.} indicates colour with additional GLCM texture statistics. \textbf{Bold} numbers highlight the best performance per column and per setting.}\centering
\label{tab:rag-egde}
\resizebox{\linewidth}{!}{
\begin{tabular}{lcccccccc}
\toprule
\textbf{RAG edge}                           & \textbf{V21}  & \textbf{PC60} & \textbf{C-Obj} & \textbf{V20}  & \textbf{PC59} & \textbf{Stuff} & \textbf{City} & \textbf{ADE}  \\ \midrule
\cellcolor[HTML]{FFCE93}CLIP\_feat          & 55.5          & 28.4          & 23.9           & 74.3          & 33.2          & 19.2           & 28.9          & 14.9          \\
\cellcolor[HTML]{FFCE93}DINO\_feat          & 55.2          & 27.9          & 25.2           & 75.3          & 32.9          & 20.2           & 29.2          & 15.4          \\
\cellcolor[HTML]{ECF4FF}C.-only             & 58.6          & 32.1          & 32.2           & 80.0          & 35.8          & 24.0           & 35.2          & 17.2          \\
\cellcolor[HTML]{ECF4FF}C. + G.             & \textbf{60.0} & \textbf{33.1} & \textbf{34.2}  & \textbf{81.0} & \textbf{36.1} & \textbf{25.5}  & \textbf{36.7} & \textbf{19.0} \\ \hline
\rowcolor[HTML]{EFEFEF} 
\cellcolor[HTML]{FFCE93}CLIP\_feat \ding{96} & 53.0          & 26.7          & 22.4           & 72.0          & 30.2          & 16.4           & 25.3          & 12.8          \\
\rowcolor[HTML]{EFEFEF} 
\cellcolor[HTML]{FFCE93}DINO\_feat \ding{96} & 53.5          & 25.4          & 21.9           & 71.8          & 31.7          & 18.9           & 26.0          & 13.0          \\
\rowcolor[HTML]{EFEFEF} 
\cellcolor[HTML]{ECF4FF}C.-only \ding{96}    & 50.4          & 27.6          & 25.3           & 74.2          & 29.9          & 18.0           & 30.4          & 10.2          \\
\rowcolor[HTML]{EFEFEF} 
\cellcolor[HTML]{ECF4FF}C. + G. \ding{96}    & \textbf{58.5}          & \textbf{32.0}          & \textbf{32.9}           & \textbf{79.9}          & \textbf{35.1}          & \textbf{23.9}           & \textbf{35.8}         & \textbf{18.2}          \\ \bottomrule
\end{tabular}}
\end{table}

\paragraph{Ablation on proposed components.} Table~\ref{tab:ablation_com} presents a component-level ablation study based on the NACLIP model, evaluating the individual and combined contributions of the Similarity Fusion module (SimFusion) and the RAG-bias mechanism. The baseline model without either component achieves an average mIoU of 39.4. Introducing SimFusion or RAG-bias individually improves the performance to 40.0 (+0.6) and 40.9 (+1.5), respectively. When both components are enabled, the model reaches the highest performance with an average mIoU of 41.2, showing consistent gains across all datasets.

These results suggest that the two components are complementary and jointly contribute to performance improvements. SimFusion enhances cross-region similarity integration, while RAG-bias introduces semantically meaningful structural bias to the attention. Notably, RAG-bias yields a greater standalone improvement than SimFusion, highlighting its stronger impact on the model’s effectiveness.

\paragraph{Ablation on RAG edge.} Table~\ref{tab:rag-egde} presents an ablation study comparing different feature types for constructing RAG edges. We evaluate four configurations: CLIP features, DINO features, colour-only input (\textit{C.-only}), and colour with additional GLCM texture statistics (\textit{C. + G.}). The top half reports results under standard conditions, while the bottom half (grey-shaded rows, marked with \ding{96}) includes colour perturbations to assess robustness.
Since our RAG is primarily constructed using low-level features, it may be susceptible to common image perturbations. To analyse model behaviour under such conditions and evaluate robustness, we apply random colour jitter using the ColorJitter function with the following parameters: brightness=0.2, contrast=0.3, saturation=0.3, and hue=0.1. This augmentation introduces appearance shifts while preserving semantic content.

We find that combining colour with GLCM texture (\textit{C.+G.}) consistently outperforms other settings under both clean and perturbed conditions. Under colour jitter, \textit{C.-only} degrades noticeably, while \textit{C.+G.} remains strong (\eg 35.8 on City, 18.2 on ADE), surpassing even CLIP and DINO. This confirms the effectiveness of integrating texture features for robust RAG edge construction.

\paragraph{Ablation on the number of neighbours.} When computing the RAG-bias (see $\mathcal{N}(i)$ in Eq. \ref{neh}), we can aggregate information from a varying number of neighbouring nodes. Table~\ref{tab:neigh} reports the results using 4 and 8 neighbours. While using 8 neighbours yields slightly better performance, the differences are marginal. This suggests that, once the RAG is constructed, our method is relatively insensitive to the number of aggregated neighbours.

\begin{table}[]
\caption{Performance comparison with different neighbour configurations. ``\#neigh." is the number of neighbours.}\centering
\label{tab:neigh}
\resizebox{\linewidth}{!}{
\begin{tabular}{ccccccccc}
\toprule
\textbf{\#neigh.} & \multicolumn{1}{c}{\textbf{V21}} & \multicolumn{1}{c}{\textbf{PC60}} & \multicolumn{1}{c}{\textbf{C-Obj}} & \multicolumn{1}{c}{\textbf{V20}} & \multicolumn{1}{c}{\textbf{PC59}} & \multicolumn{1}{c}{\textbf{Stuff}} & \multicolumn{1}{c}{\textbf{City}} & \multicolumn{1}{c}{\textbf{ADE}} \\ \midrule
4            & 60.1                             & 33.0                              & 34.2                               & 81.0                             & 35.8                              & 25.6                               & 36.5                              & 19.0                             \\
8            & 60.2                             & 33.4                              & 34.4                               & 81.0                             & 36.2                              & 25.8                               & 36.9                              & 19.2                             \\ \bottomrule
\end{tabular}}
\end{table}

\paragraph{Ablation on patch size and image size.} Since our proposed superpixel-aligned patch encoding method computes representations based on the superpixel regions within each patch, both the patch size and input image resolution may affect the model performance. As shown in Table~\ref{tab:patch_resolution}, using smaller patch sizes (\eg B/16 \textit{vs.} B/32) and higher image resolutions (\eg 336 \textit{vs.} 224) consistently leads to better performance across all benchmarks.
In particular, the best results (an average mIoU of 40.9) are achieved with the B/16 patch size and 336 image resolution setting. These results suggest that finer spatial granularity in the patch-level representation helps better capture region-boundary alignment with superpixels, enhancing segmentation quality.

\begin{table}
\centering
\caption{Effect of patch size and image resolution on performance. The results are reported on the NACLIP model.}
\resizebox{\linewidth}{!}{
\begin{tabular}{llcccccccc}
\toprule
\textbf{Patch} & \textbf{Img} & \textbf{V21}  & \textbf{PC60} & \textbf{C-Obj} & \textbf{V20}  & \textbf{PC59} & \textbf{Stuff} & \textbf{City} & \textbf{ADE} \\
% \textbf{Size} & \textbf{Size} &     &      &       &     &      &       &      &     \\
\midrule
B/16 & 336 & \textbf{60.2} & \textbf{33.4} & \textbf{34.4} & \textbf{81.0} & \textbf{36.2} & \textbf{25.8} & \textbf{36.9} & \textbf{19.2} \\
B/32 & 336 & 57.2 & 30.0 & 31.2 & 78.7 & 34.3 & 22.7 & 32.5 & 16.4 \\
B/16 & 224 & 58.5 & 31.2 & 30.9 & 79.2 & 35.0 & 23.6 & 34.0 & 16.1 \\
B/32 & 224 & 55.2 & 28.0 & 29.6 & 75.0 & 31.7 & 20.0 & 30.6 & 13.6 \\
\bottomrule
\end{tabular}}
\label{tab:patch_resolution}
\end{table}

\paragraph{Ablation on different superpixel methods.} In Table~\ref{tab:superpixel_comparison}, we compare different superpixel segmentation methods for RAG construction, including SLIC, Watershed, and Felzenszwalb. Among them, SLIC achieves the best overall performance, while Felzenszwalb performs the worst across all benchmarks.
We attribute the weaker performance of Felzenszwalb to its irregular and region-driven segmentation outputs, which often deviate from the grid-like patch structure used in our superpixel-to-patch encoding. In contrast, SLIC produces more compact and uniformly shaped superpixels that align better with patch boundaries, making it more compatible with our proposed encoding strategy.

\begin{table}[]
\centering
\caption{Comparison of different superpixel segmentation methods for RAG construction. Best performance in \textbf{bold}. A discussion on using masks generated by the Segment Anything Model (SAM)~\cite{kirillov2023segment} is provided in Supplementary Material Section~\ref{s5}.}
\resizebox{\linewidth}{!}{
\begin{tabular}{lcccccccc}
\toprule
\textbf{Method} & \textbf{V21}  & \textbf{PC60} & \textbf{C-Obj} & \textbf{V20}  & \textbf{PC59} & \textbf{Stuff} & \textbf{City} & \textbf{ADE}  \\ \midrule
SLIC            & \textbf{60.2} & \textbf{33.4} & \textbf{34.4}  & \textbf{81.0} & \textbf{36.2} & \textbf{25.8}  & \textbf{36.9} & \textbf{19.2} \\
Watershed       & 58.2          & 32.8          & 33.2           & 80.2          & 35.3          & 23.2           & 34.9          & 18.2          \\
Felzenszwalb    & 55.2          & 30.1          & 28.9           & 76.9          & 33.0          & 22.5           & 32.9          & 14.5          \\ \bottomrule
\end{tabular}}
\label{tab:superpixel_comparison}
\end{table}

\paragraph{Ablation on $\alpha$ in similarity fusion.} Table~\ref{tab:sim} shows the performance of our model under different weighting parameter $\alpha$ in the Similarity Fusion module. We observe that the module with any $\alpha>0$ consistently improves performance over the baseline (w/o), suggesting the effectiveness of combining the original similarity matrix with its smoothed counterpart. This fusion helps suppress noisy or unreliable similarity signals, leading to more robust region aggregation. The best performance is achieved at $\alpha=0.6$, which is used as the default setting in all experiments.

\paragraph{Generalisation analysis.} We provide an extensive analysis of our model's generalisation capabilities in Section~\ref{s2} of the Supplementary Material, evaluating its performance on challenging transformations such as overexposure, underexposure, grayscale, style transfer, and texture destruction. In addition, we assess its zero-shot performance on a remote sensing dataset, with results reported in the Supplementary Material in Table~\ref{table:postdam} and Fig.~\ref{fig:postdam}.

\vspace{-1em}
\paragraph{Failure case analysis.} Our method's primary failure cases occur in underexposed conditions, where the loss of fine-grained local details causes our attention bias to be mis-weighted. For a detailed analysis of these failure cases, please refer to Section~\ref{s3} in the Supplementary Material.

\begin{table}
\footnotesize
\centering
\caption{Efficiency analysis of our method. Comparison of inference speed (FPS) and computational cost (FLOPs) on different baseline models.}
\begin{tabular}{lcc}
\toprule
\textbf{Method} & \textbf{FPS ($\Delta$)} & \textbf{FLOPs ($\Delta$)} \\
\midrule
CLIP / + Ours & 72.5 $\rightarrow$ 71.1\ (-1.4) & 41.7 $\rightarrow$ $\approx$41.7\ (+0.0) \\
SCLIP / + Ours & 68.8 $\rightarrow$ 66.9\ (-1.9) & 44.2 $\rightarrow$ $\approx$44.2\ (+0.0) \\
ProxyCLIP / + Ours & 52.9 $\rightarrow$ 51.5\ (-1.4) & 81.1 $\rightarrow$ $\approx$81.1\ (+0.0) \\
\bottomrule
\label{fig:cost_time}
\end{tabular}
\end{table}

\begin{table}[]\centering
\caption{Performance comparison under different $\alpha$ values on SimFusion module based on the NACLIP model.}
\label{tab:sim}
\resizebox{\linewidth}{!}{
\begin{tabular}{l|cccccccc}
\toprule
\textbf{$\alpha$} & \textbf{V21}  & \textbf{PC60} & \textbf{C-Obj} & \textbf{V20}  & \textbf{PC59} & \textbf{Stuff} & \textbf{City} & \textbf{ADE}  \\ \midrule
\rowcolor[HTML]{EFEFEF} 
w/o               & 58.9          & 32.2          & 33.2           & 79.7          & 35.2          & 23.3           & 35.5          & 17.4          \\ \hline
0.1               & 58.9          & 32.2          & 33.2           & 79.7          & 35.2          & 23.3           & 35.5          & 17.6          \\
0.2               & 59.0          & 32.5          & 33.4           & 79.9          & 35.3          & 23.5           & 35.6          & 17.7          \\
0.5               & 59.2          & 32.5          & \textbf{33.6}  & 80.3          & 35.8          & \textbf{24.5}  & 35.9          & 18.0          \\
\textbf{0.6}      & \textbf{59.4} & \textbf{32.6} & 33.5           & \textbf{80.5} & \textbf{35.9} & 24.2           & \textbf{36.0} & \textbf{18.1} \\
0.7               & 58.9          & 32.0          & 33.1           & 80.3          & 35.8          & 24.1           & 35.9          & 18.0          \\ \bottomrule
\end{tabular}}
\end{table}

% \paragraph{Qualitative results.} The Fig.~\ref{fig:vis} presents qualitative comparisons between our method and CLIPtrace. As shown, our method produces more coherent segmentation results, particularly in challenging regions such as object boundaries and small structures. The highlighted areas (in red boxes) demonstrate that our model better preserves local consistency, effectively reducing fragmented or noisy predictions commonly observed in the CLIPtrace outputs. This supports the effectiveness of our proposed components in refining semantic segmentation at a finer granularity.

\begin{figure}[!h]
    \centering
    \includegraphics[width=\linewidth]{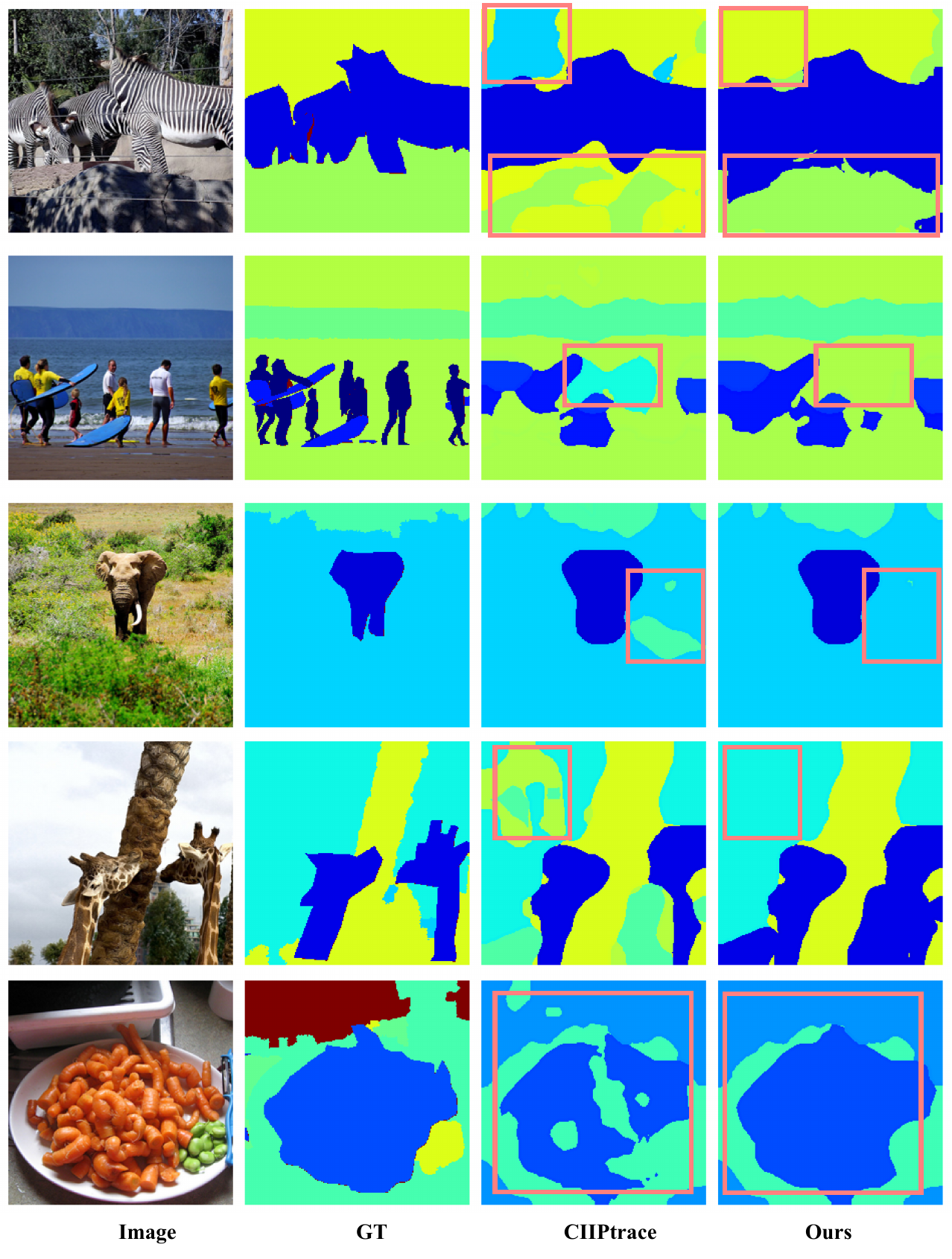} % 预定义占位图片
    \caption{The qualitative results of our method. For more challenging cases, such as grayscale and stylised images (\eg oil paintings), please refer to Figs.~\ref{fig:lighting}, \ref{fig:destruction}, and \ref{fig:domain_shift} in the Supplementary Materials.}% between our method and CLIPtrace~\cite{cliptrace}.}
    \label{fig:vis}
\end{figure}

\vspace{-1em}
\paragraph{Qualitative results.} In Fig.~\ref{fig:vis} we present qualitative comparisons between our method and CLIPtrace. As illustrated, our approach consistently yields more coherent and accurate segmentation results, especially in challenging regions such as object boundaries, fine-grained structures, and texture-rich areas. The highlighted areas (in red boxes) emphasise our model’s ability to preserve local consistency and object integrity, reducing the fragmented or noisy predictions that are often observed in the CLIPtrace outputs. These qualitative observations underscore the effectiveness of our proposed components in enhancing performance at finer levels of granularity.

\vspace{-1em}
\paragraph{Computational cost.} As shown in Table \ref{fig:cost_time}, our method is computationally efficient. When integrated with various baselines, it introduces no additional FLOPs while only causing a negligible decrease in inference speed (FPS).
 
\section{Conclusion}

In this work, we proposed a new feature rectification approach for training-free open-vocabulary semantic segmentation. Our method leverages a Region Adjacency Graph (RAG) to refine visual patch embeddings and address local inconsistency in CLIP-based models. Specifically, we introduced a RAG-bias to guide attention toward semantically relevant regions, and a Similarity Fusion module to better align visual patches with textual categories.
Extensive experimental analysis showed the effectiveness and generalisability of our approach, presenting consistent improvements across multiple datasets without additional training. Ablation studies further highlighted the importance of neighbourhood design and RAG construction, providing insights into utilising low-level priors for semantic refinement.

\clearpage
\section*{Acknowledgements}
This project is partially supported by an Amazon Research Award.
Qiming Huang is supported by the China Scholarship Council (Grant No. 202408060321).
The computations in this research were performed using the Baskerville Tier 2 HPC service. Baskerville was funded by the EPSRC and UKRI through the World Class Labs scheme (EP\textbackslash T022221\textbackslash1) and the Digital Research Infrastructure programme (EP\textbackslash W032244\textbackslash1) and is operated by Advanced Research Computing at the University of Birmingham.

\section*{Appendix}

\renewcommand{\thesection}{S\arabic{section}}
\renewcommand{\thefigure}{S\arabic{figure}}
\renewcommand{\thetable}{S\arabic{table}}
\setcounter{figure}{0}
\setcounter{table}{0}
\setcounter{section}{0}

\section{More ablation study for hyperparameters}\label{s1}

To investigate the impact of key hyperparameters on our model's performance, we conducted a series of ablation studies. The experiments focused on the parameters of the Simple Linear Iterative Clustering (SLIC) algorithm and the selection of features from the Grey-Level Co-occurrence Matrix (GLCM).

% As depicted in Fig. \ref{fig:slic}, we performed a grid search to optimise the number of segments and the compactness value for the SLIC algorithm. The results on the COCO-Stuff-171 validation set indicate that the peak performance, an mIoU of 25.47, was achieved when the number of segments was set to 300 and compactness to 10. This finding underscores the sensitivity of the model to the SLIC parameterization.

\begin{figure}[!h]
    \centering
    \includegraphics[width=1.2\linewidth]{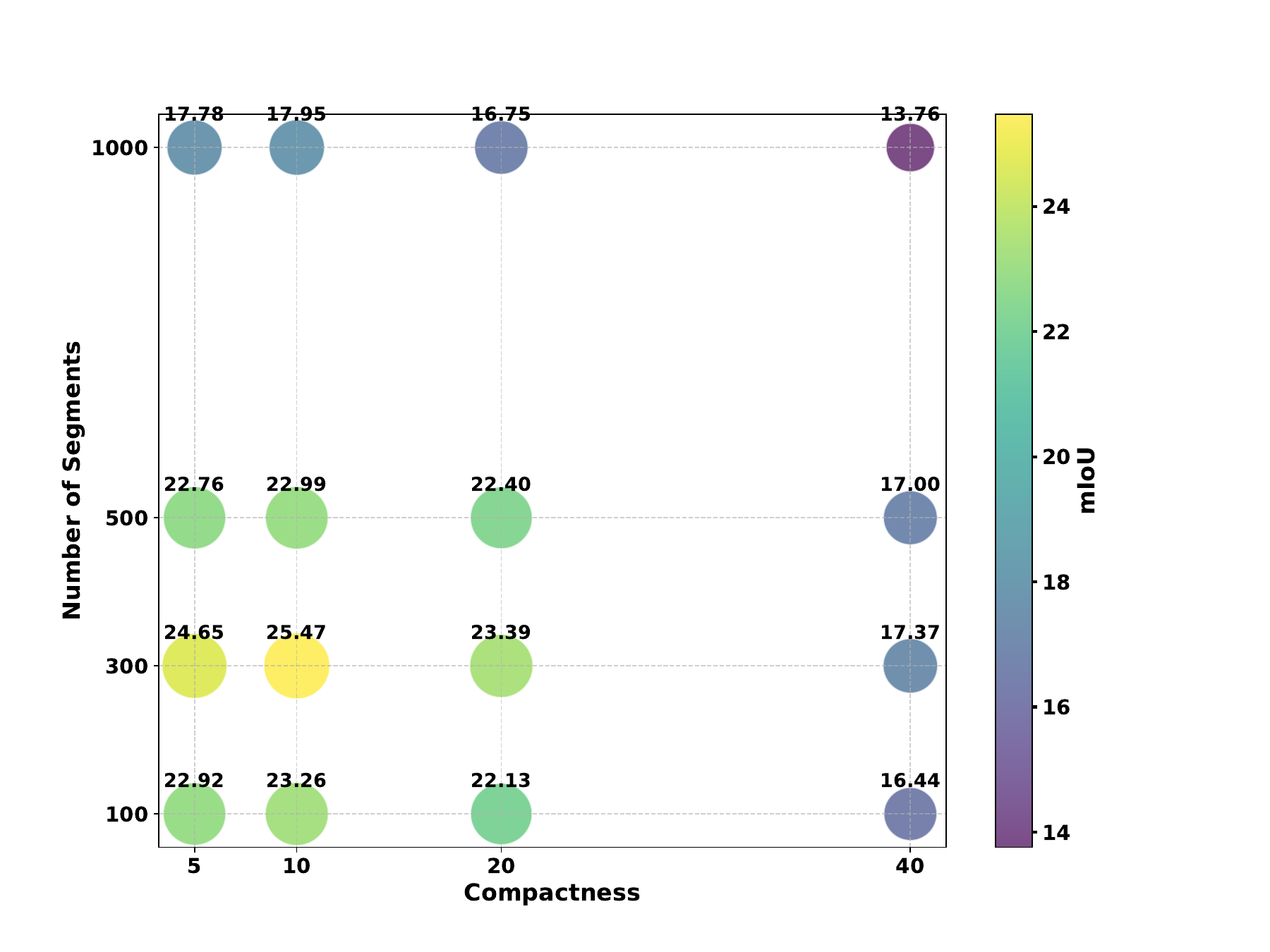}    
    \caption{Ablation study on the number of segments (n\_segments) and compactness for the SLIC algorithm. Results are reported as mIoU on the COCO-Stuff-171 validation set. The baseline model is NACLIP.}
    \label{fig:slic}
\end{figure}

\begin{figure*}[t]
    \centering
    \includegraphics[width=\linewidth]{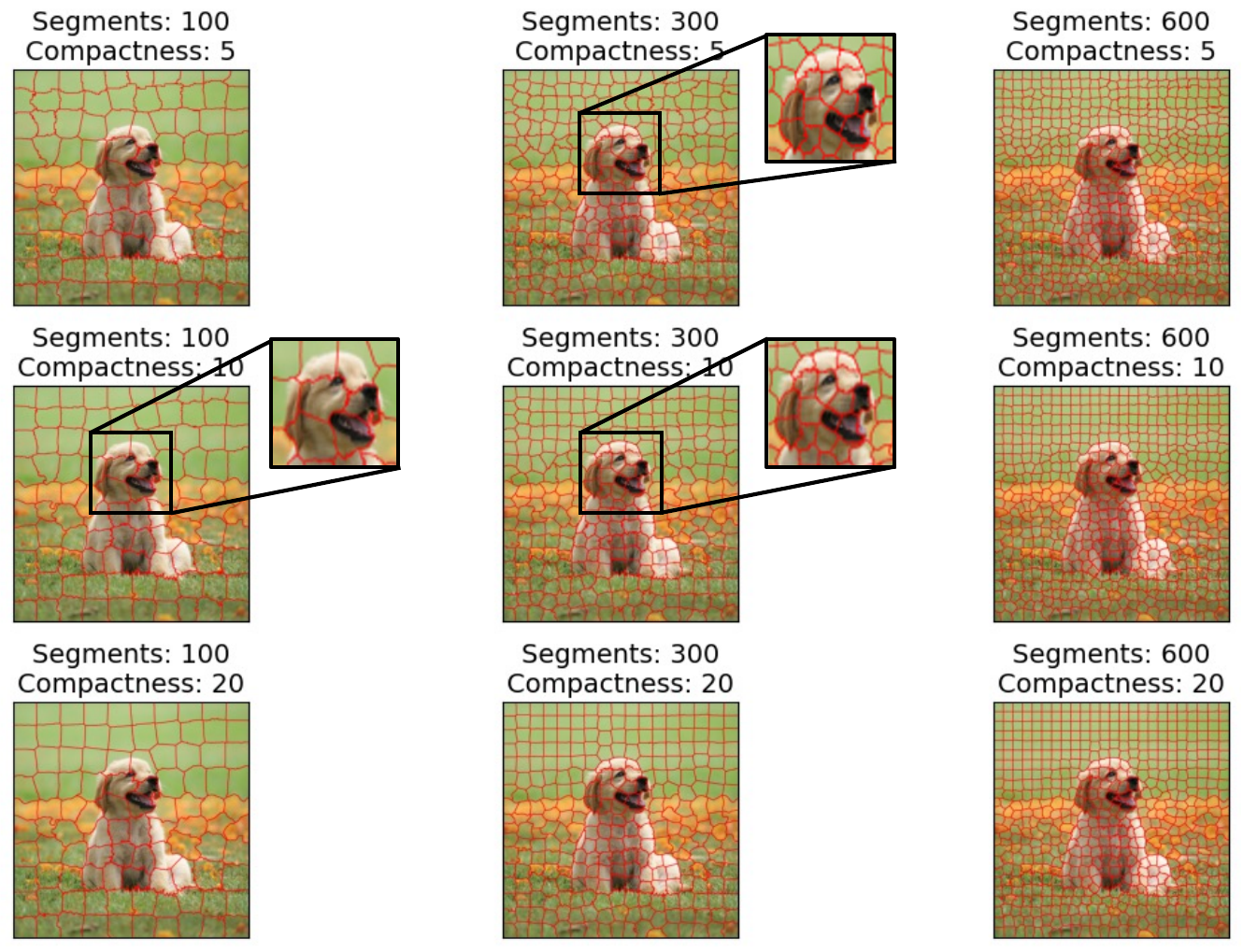}
    \caption{Comparison of SLIC segmentation results across different combinations of Segments (number of superpixels) and Compactness.}
    \label{fig:slic_}
\end{figure*}

We performed a grid search to optimise the number of segments and the compactness for the SLIC algorithm, with the quantitative results shown in Fig. \ref{fig:slic}. Different hyperparameters of SLIC produce different region proposal results, as visualised in Fig. \ref{fig:slic_}. The number of segments directly controls the scale of the superpixels; increasing this value results in finer, more numerous regions. The compactness parameter manages the trade-off between spatial proximity and colour similarity; a lower value allows superpixels to conform more closely to image textures and edges, while a higher value produces more uniform, regularly shaped regions. Our analysis indicates that the optimal balance for this task was achieved with 300 segments and a compactness of 10, yielding a peak mIoU of 25.47. 

\begin{figure}
    \centering
    \includegraphics[width=\linewidth]{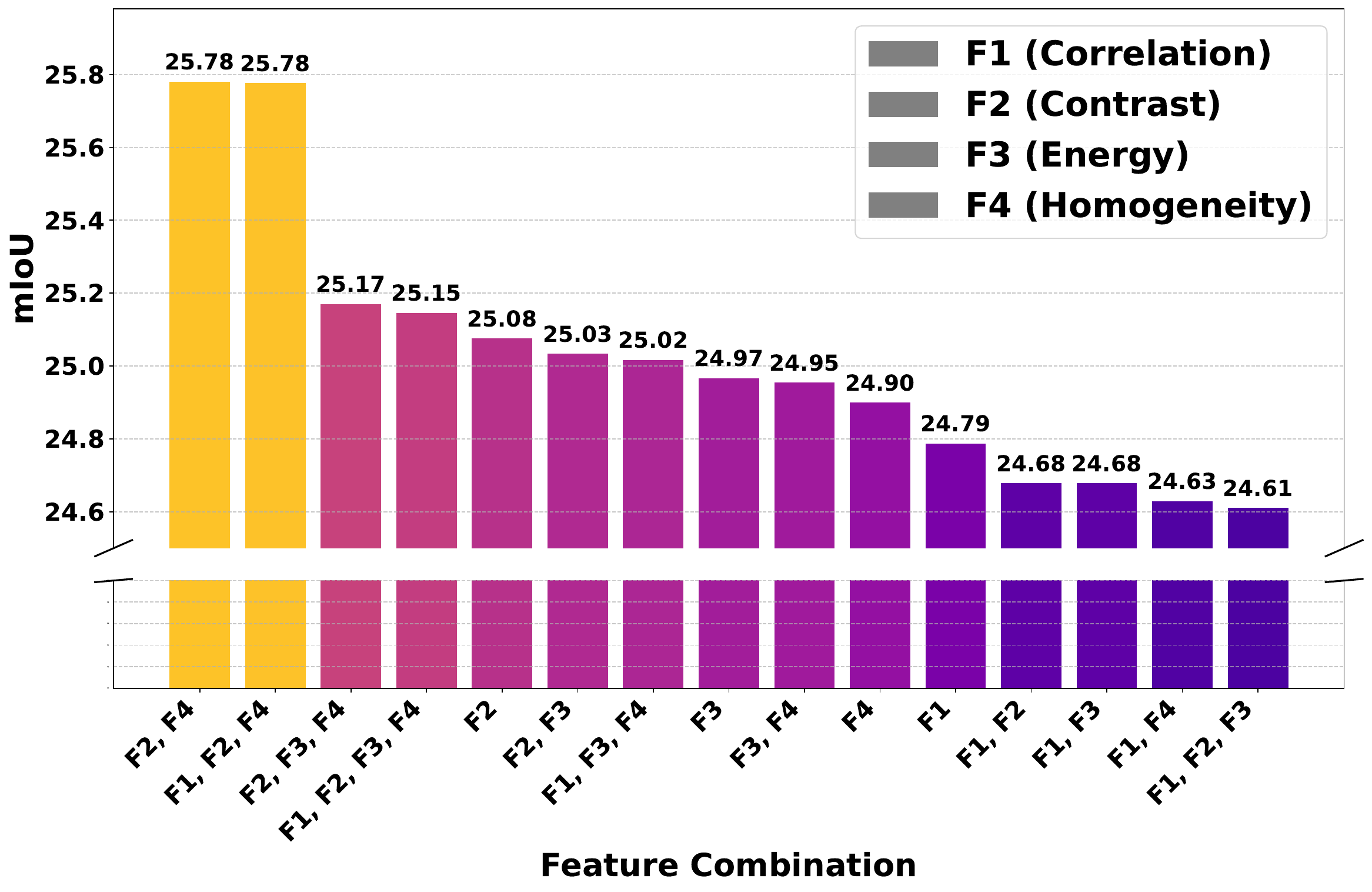}    
    \caption{The impact of different SLIC feature combinations on the performance of the NACLIP baseline. The experiment was conducted on the COCO-Stuff-171 validation set with baseline NACLIP.}
    \label{fig:glcm}
\end{figure}

Furthermore, we analysed the contribution of different combinations of four GLCM texture features: Correlation (F1), Contrast (F2), Energy (F3), and Homogeneity (F4). The results, presented in Fig. \ref{fig:glcm}, reveal that the combination of Contrast (F2) and Homogeneity (F4) yielded the highest mIoU of 25.78. Notably, this two-feature subset outperformed the combination of all four features, highlighting that an appropriate selection of features is more effective than using them all.

% \begin{figure*}
%     \centering
%     \includegraphics[width=\linewidth]{src/vis_slic.pdf}
%     \caption{A comparison of SLIC segmentation results across different combinations of Segments (number of superpixels) and Compactness.}
%     \label{fig:slic_}
% \end{figure*}

% \twocolumn[{%
% \renewcommand\twocolumn[1][]{#1}%
% \maketitle
% \begin{center}
%     \includegraphics[width=\linewidth]{src/supp/vis_slic.pdf}
%     \captionof{figure}{Comparison of SLIC segmentation results across different combinations of Segments (number of superpixels) and Compactness.}
%     \label{fig:slic_}
% \end{center}%
% }]

\section{Generalisation analysis}\label{s2}

To further evaluate the generalisation capabilities of our proposed method, we conduct a series of analyses under various challenging conditions, including common image corruptions, domain shifts, and zero-shot segmentation on extra domain-related remote sensing datasets postdam\footnote{Isprs potsdam dataset on kaggle. Available online: https://www.kaggle.com/datasets/jahidhasan66/isprs-potsdam}.

\paragraph{Robustness to common image corruptions.} We first assess the model's resilience to common visual perturbations that degrade image quality. Table \ref{table:per} quantitatively measures the performance changes of our method built upon NACLIP baseline under four conditions: overexposure, underexposure, grayscale conversion, and texture destruction (via Gaussian blur). For overexposure, we used a brightness factor of 1.8 to significantly increase the luminosity, causing highlights to become washed out. For underexposure, we used a brightness factor of 0.4 to decrease the luminosity, obscuring details in shadows. For texture destruction, we used a kernel size of $(9, 9)$ pixels and $\sigma=5$ to create a significant and noticeable blurring effect that effectively destroys surface textures. Our method maintains reasonable performance in most cases, but significantly reduces when facing strong low exposure.

\begin{figure*}[t]
    \centering
    \includegraphics[width=\linewidth]{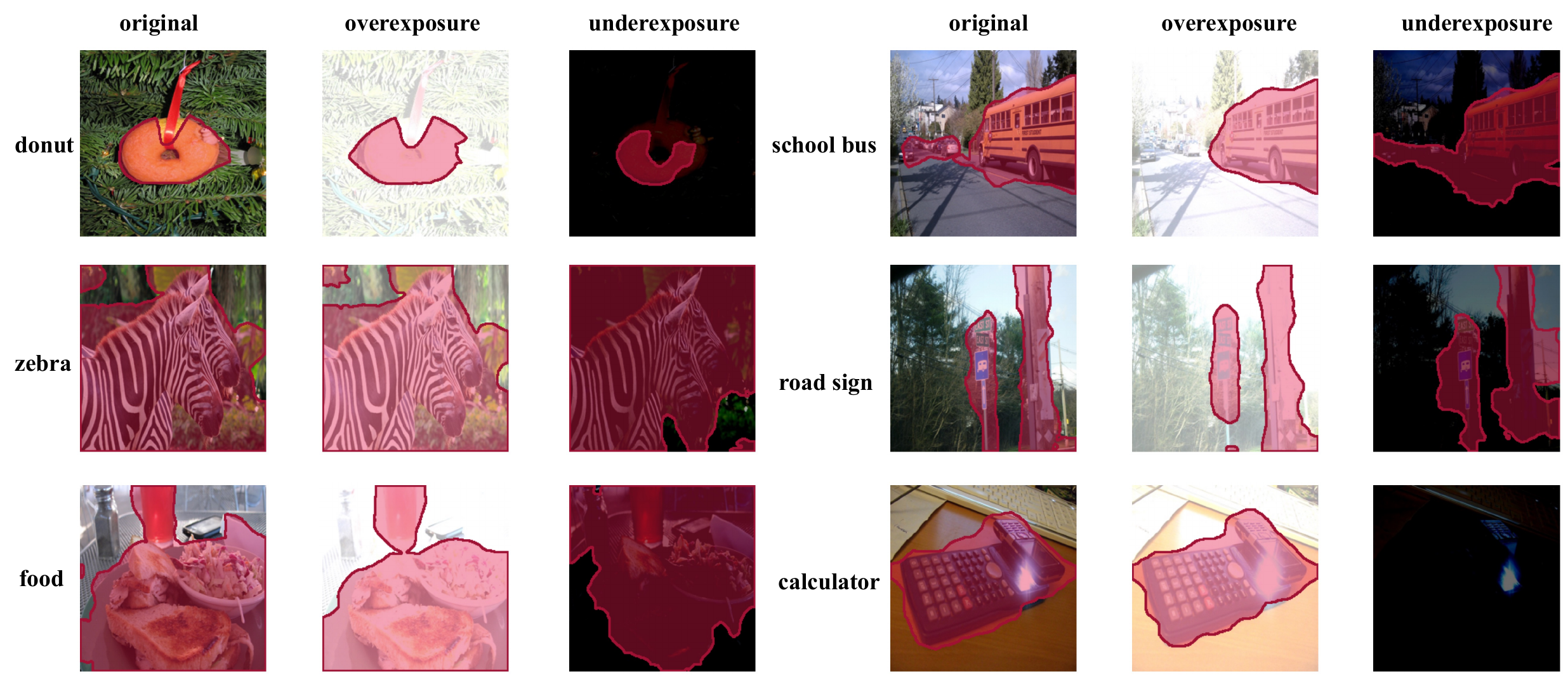}
    \caption{The effect of overexposure and underexposure on segmentation performance.}
    \label{fig:lighting}
\end{figure*}

To better understand this, we provide a visualisation of the segmentation results under these conditions. \textbf{Effect of Lighting.} As shown in Fig. \ref{fig:lighting}, our model performs well under extreme lighting changes. Despite significant information loss in the bright, washed-out areas of overexposed images or the dark, detail-lacking regions of underexposed images, our model consistently generates reasonable segmentation masks for objects like `doughnuts', `zebras'.

% \twocolumn[{%
% \renewcommand\twocolumn[1][]{#1}%
% \maketitle
% \begin{center}\label{fs4}
%     \includegraphics[width=\linewidth]{src/supp/light_vis.pdf}
%     \captionof{figure}{The effect of overexposure and underexposure on segmentation performance.}
%     \label{fig:lighting}
% \end{center}%
% }]

% \begin{figure}
%     \centering
%     \includegraphics[width=\linewidth]{src/light_vis.pdf}
%     \caption{The effect of overexposure and underexposure on segmentation performance.}
%     \label{fig:lighting}
% \end{figure}

\begin{figure*}
    \centering
     \includegraphics[width=\linewidth]{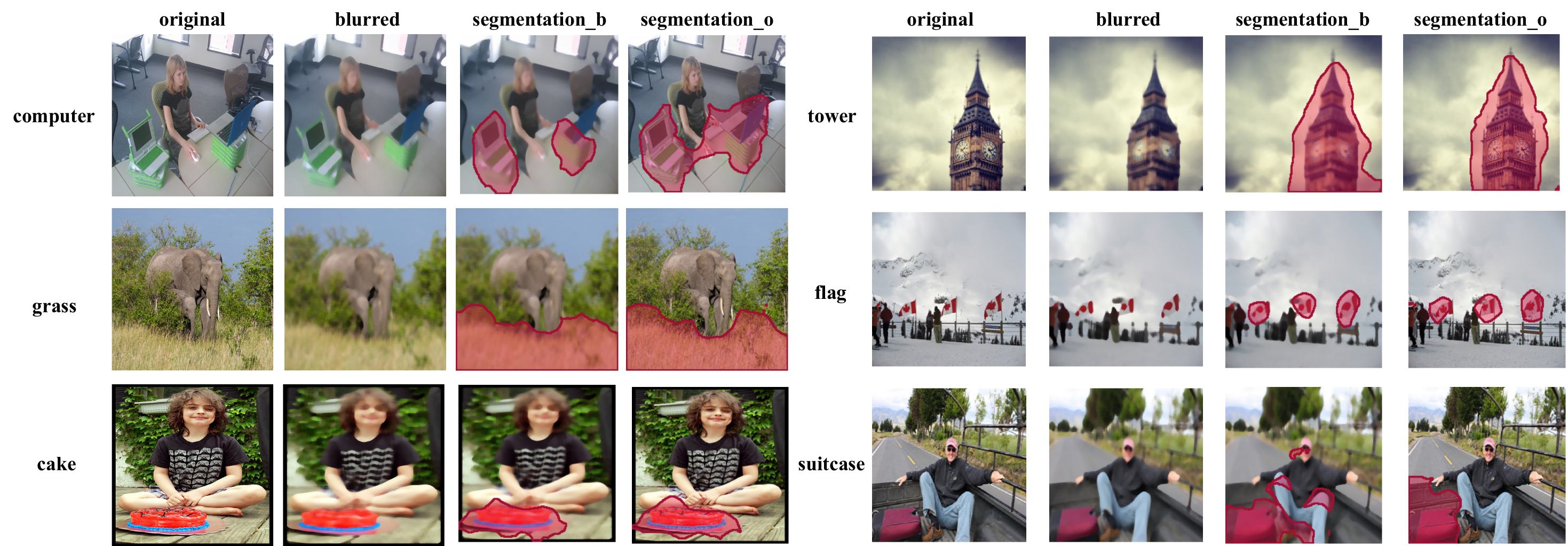}
    \captionof{figure}{The effect of texture destruction (Gaussian blur) on segmentation performance.}
    \label{fig:destruction}
\end{figure*}

\textbf{Effect of texture destruction.} To simulate the loss of fine-grained details and high-frequency textures, we applied a Gaussian blur filter. This process involves convolving the image with a Gaussian kernel. The intensity of the blur is controlled by the kernel size and the standard deviation ($\sigma$). In our experiments, we used a kernel size of $(9, 9)$ pixels and $\sigma=5$ to create a significant and noticeable blurring effect that effectively destroys surface textures. The visualisation results are shown in Fig.~\ref{fig:destruction}.

\begin{table}[]
\centering
\caption{The performance drops of our method under different cases. We use NACLIP as the baseline.}
\begin{tabular}{l|llll}
\toprule
                    & V20 & Stuff & PC59 & ADE \\ \midrule
overexposure        & -2.1    & -1.8      & -2.5     & -1.5    \\
underexposure       & -8.5    & -10.8      &  -7.5    & -4.5    \\
grayscale           & -1.1    & -1.5      & -2.1     &  -1.2   \\
texture destruction & -1.8    & -2.3      &  -2.2    &  -1.8   \\ \bottomrule
\end{tabular}
\label{table:per}
\end{table}

\begin{figure}
    \centering
    \includegraphics[width=\linewidth]{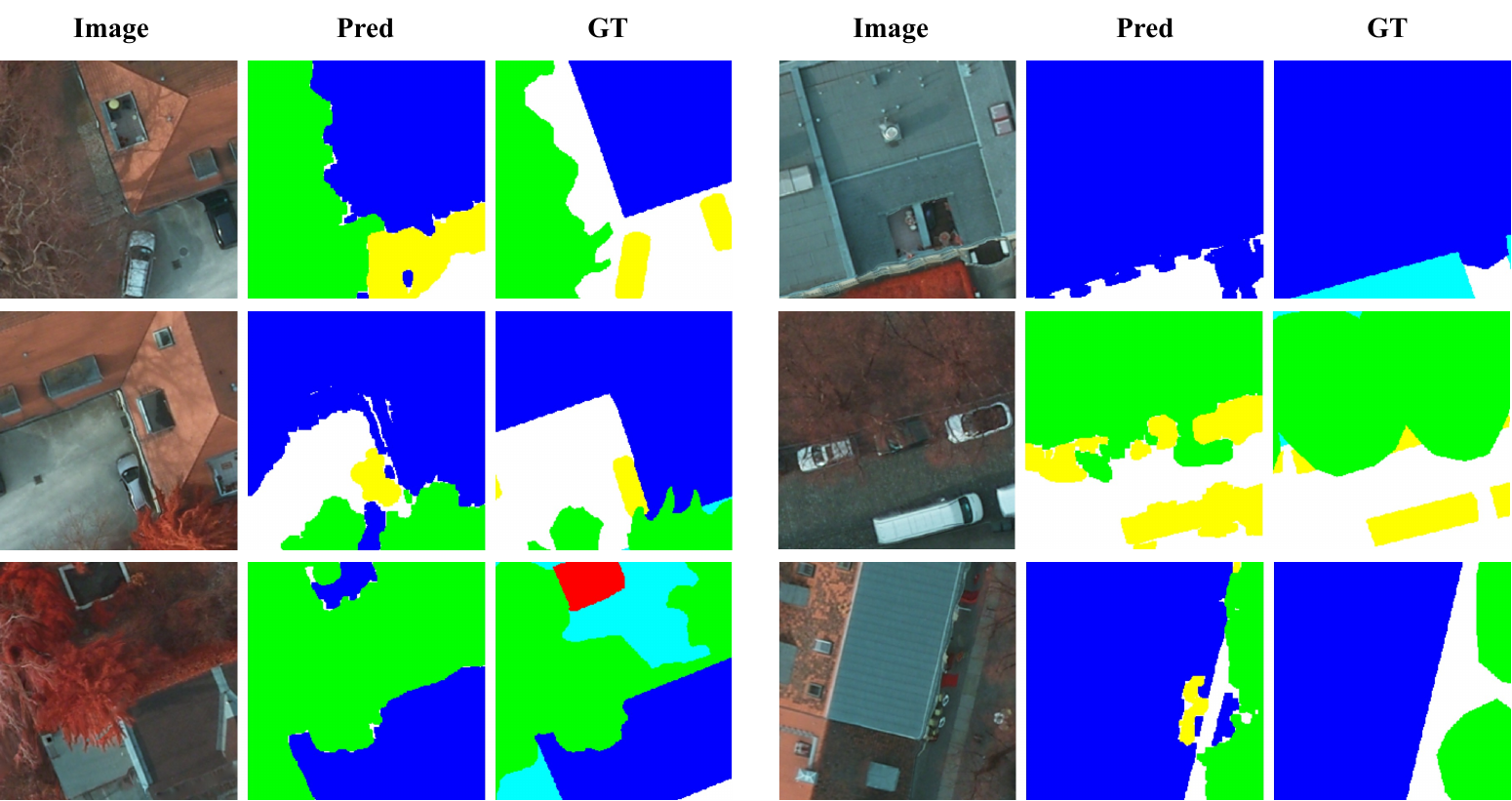}
    \caption{Visualisation results on the Potsdam dataset of our method built upon NACLIP.}
    \label{fig:postdam}
\end{figure}

\begin{table}[]
\centering
\caption{Comparison of Zero-Shot Performance on the Potsdam Remote Sensing Datasets. Results are reported on mIoU}
\begin{tabular}{ll}
\toprule
                   & Potsdam  \\ \midrule
NACLIP             & 28.6         \\
NACLIP + Ours      & 30.4         \\ \hline
ClipSurgery        & 30.2         \\ 
ClipSurgery + Ours & 32.1         \\ \bottomrule
\end{tabular}
\label{table:postdam}
\end{table}

\begin{figure*}
    \centering
    \includegraphics[width=\linewidth]{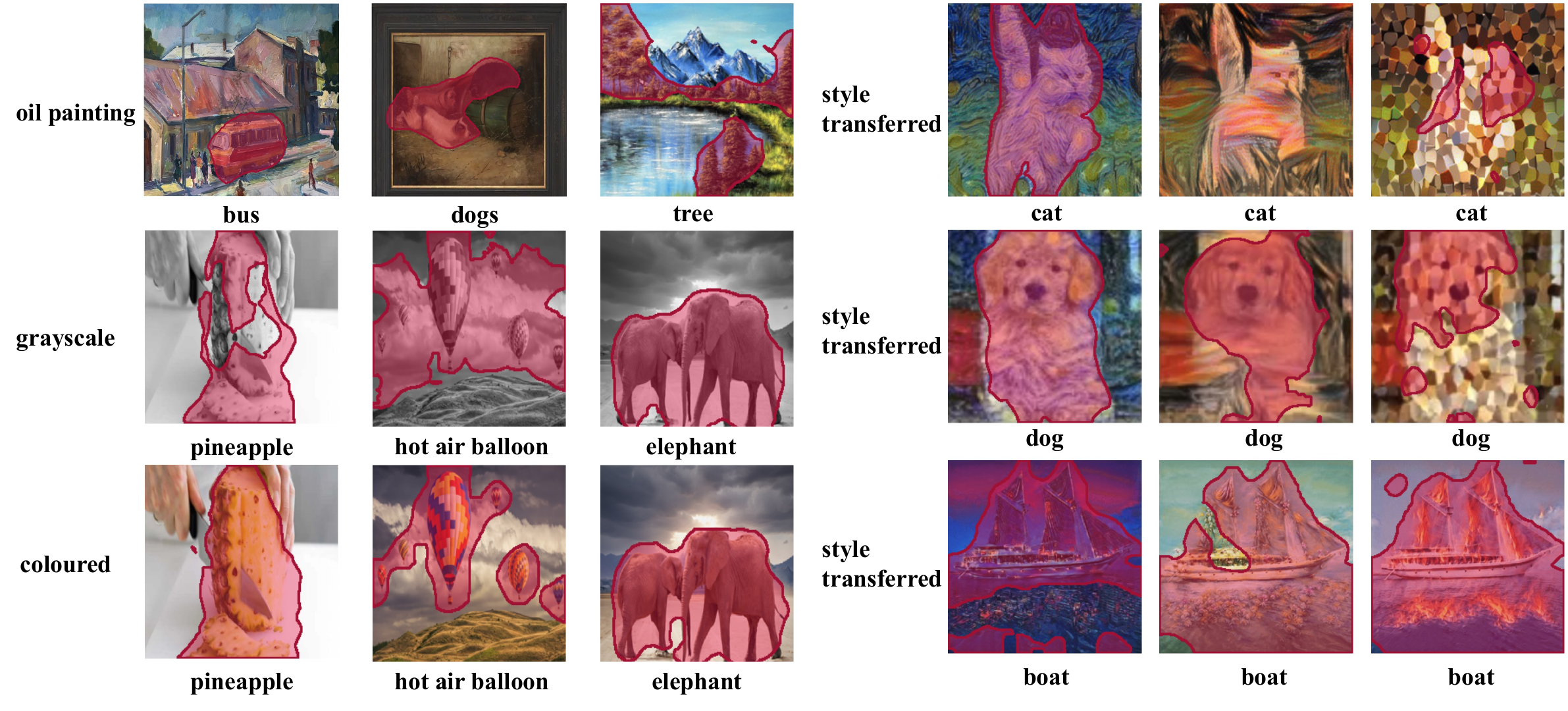}
    \captionof{figure}{The effect of domain shift on segmentation performance.}
    \label{fig:domain_shift}
\end{figure*}

\textbf{Analysis of domain shift.} We further investigate the model's ability to generalise across different visual domains, a critical aspect of real-world applications. Fig.~\ref{fig:domain_shift} showcases the segmentation performance on images that have undergone significant style and domain shifts. We test on artistic renderings (\eg oil painting), images with altered colour schemes (grayscale vs. coloured), and various other style-transferred examples. The model consistently produces precise segmentations for objects like 'dogs', 'pineapples', and 'boats' across these diverse visual styles.

\textbf{Zero-Shot generalisation to Postdam remote sensing dataset.} Additionally, we evaluate our method's zero-shot performance on a completely unseen and specialised domain: the Potsdam remote sensing dataset. As reported in Table \ref{table:postdam}, when our module is integrated with existing baselines (NACLIP and ClipSurgery), it yields substantial improvements in mean Intersection over Union (mIoU). The Visualisation result is shown in Fig. \ref{fig:postdam}.

% \twocolumn[{%
% \renewcommand\twocolumn[1][]{#1}%
% \maketitle
% \begin{center}
%     \includegraphics[width=\linewidth]{src/challenge_vis.pdf}
%     \captionof{figure}{The effect of domain shift on segmentation performance.}
%     \label{fig:domain_shift}
% \end{center}%
% }]

% \twocolumn[{%
% \renewcommand\twocolumn[1][]{#1}%
% \maketitle
% \begin{center}
%     \includegraphics[width=\linewidth]{src/supp/failure_case.pdf}
%     \captionof{figure}{Illustration of challenging scenarios leading to segmentation failures. The cases, from left to right, include: (a) an object that is too small to be accurately detected (faucet); (b) severe underexposure resulting in loss of detail; (c) disordered color and texture, making boundaries ambiguous; (d) an object camouflaged against a similar background; and (e) a scene with excessive colour complexity and numerous small details.}
%     \label{fig:failures}
% \end{center}%
% }]

\begin{figure*}[t]
    \centering
    \includegraphics[width=\linewidth]{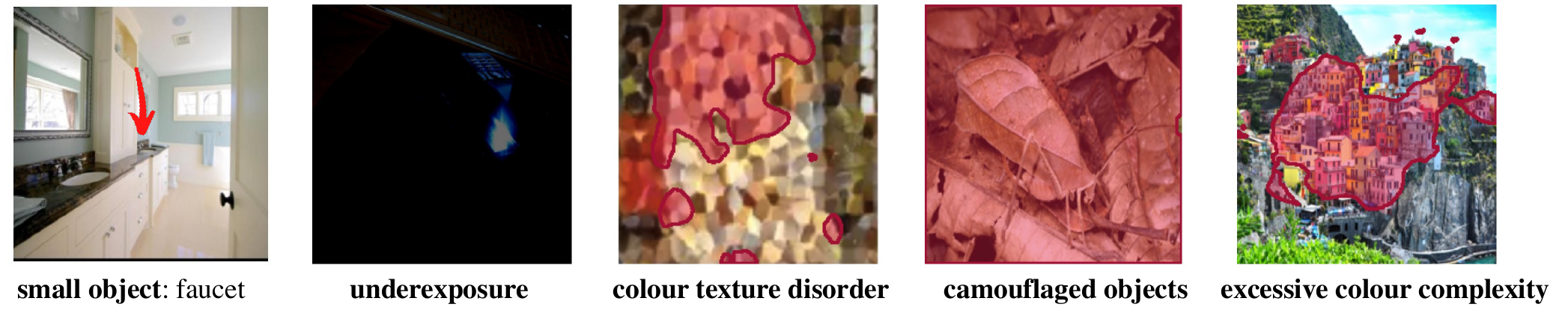}
    \caption{Illustration of challenging scenarios leading to segmentation failures. The cases, from left to right, include: (a) an object that is too small to be accurately detected (faucet); (b) severe underexposure resulting in loss of detail; (c) disordered color and texture, making boundaries ambiguous; (d) an object camouflaged against a similar background; and (e) a scene with excessive colour complexity and numerous small details.}
    \label{fig:failures}
\end{figure*}

\section{Analysis of failure cases}\label{s3} 

We analysed the failure cases of our method to better understand its limitations. Fig. \ref{fig:failures} illustrates five representative scenarios where the segmentation performance is compromised: \textbf{Small Object Insensitivity.} Our approach relies on an initial superpixel segmentation. Consequently, objects that are exceptionally small, such as the `faucet', may be smaller than the generated superpixels and are incorrectly absorbed into larger background regions. This prevents them from being represented as distinct nodes in the region adjacency graph. \textbf{Extreme Lighting Conditions.} In cases of severe underexposure, the lack of sufficient colour and brightness information cripples the feature extraction process. Both SLIC and GLCM features become unreliable, leading to a near-complete failure to identify any objects. \textbf{Ambiguous Boundaries and Camouflage.} The model's performance degrades when there is no clear distinction between foreground and background. This occurs in scenes with chaotic colour and texture, where boundaries are inherently ambiguous, and in cases of camouflage, where the object's texture features are nearly identical to the background's. \textbf{Excessive Scene Complexity.} Our method can be challenged by scenes containing an overwhelming density of small, intricate details. The high frequency of colour and texture changes results in an overly fragmented superpixel map and a highly complex region graph, which hinders the effective propagation and feature rectification.

\section{Visualisation comparison with ClipSurgery}\label{s4}

\begin{figure*}
    \centering
    \includegraphics[width=\linewidth]{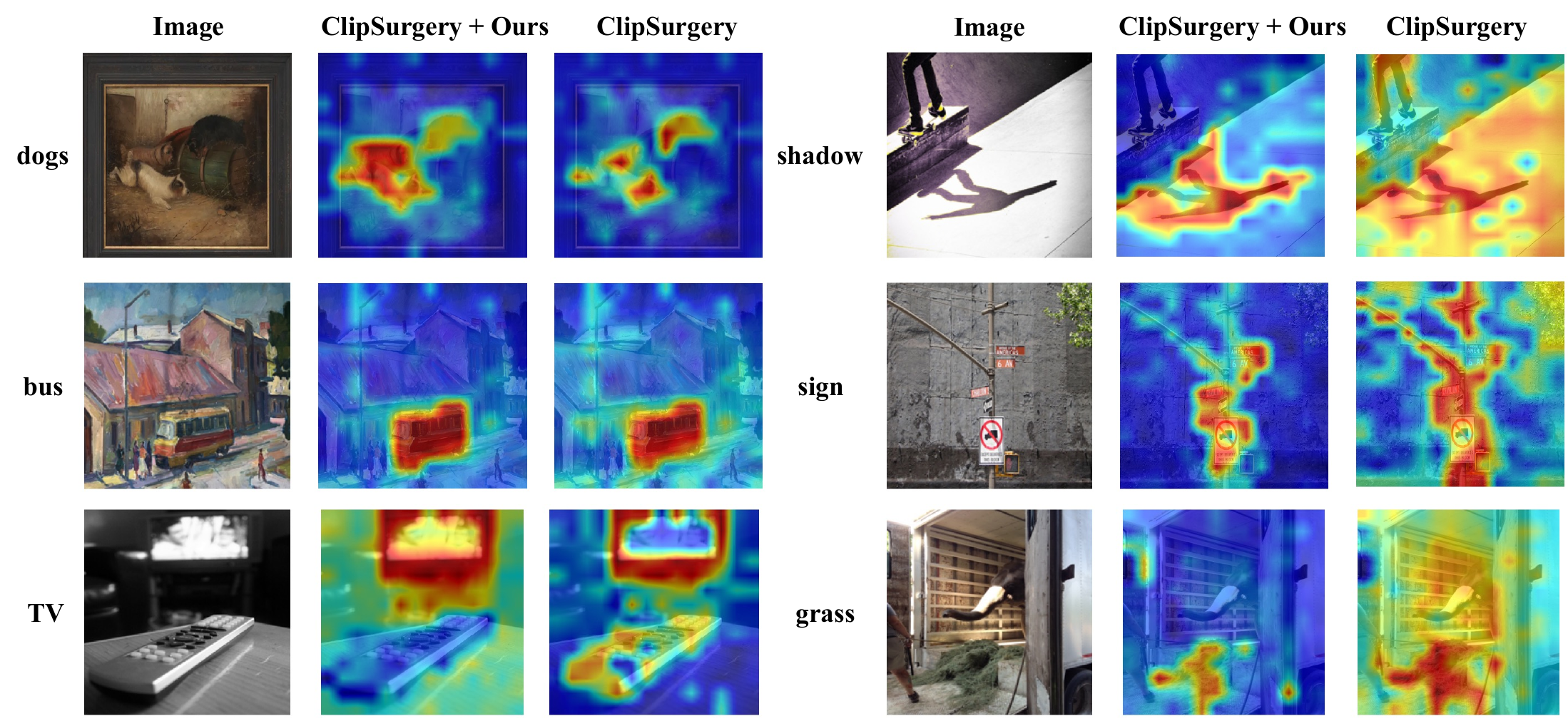}
    \caption{Comparison of model attention visualisation results with ClipSurgery and ClipSurgery + Ours}
    \label{fig:clipsurgery}
\end{figure*}

Fig. \ref{fig:clipsurgery} visualises the qualitative impact of our method when applied to ClipSurgery. It is evident that our approach refines the model's attention mechanism. The baseline ClipSurgery model, while effective, often produces coarse and noisy attention maps that fail to precisely localise the target object (\eg 'bus', 'grass'). By incorporating our structure-aware feature rectification using region adjacency graphs, the resulting attention becomes more focused and clean. Our method successfully prunes background noise and sharpens the activation to align with true object boundaries.

% \begin{figure*}
%     \centering
%     \includegraphics[width=\linewidth]{src/failure_case.pdf}
%     \caption{ Illustration of challenging scenarios leading to segmentation failures. The cases, from left to right, include: (a) an object that is too small to be accurately detected (faucet); (b) severe underexposure resulting in loss of detail; (c) disordered color and texture, making boundaries ambiguous; (d) an object camouflaged against a similar background; and (e) a scene with excessive colour complexity and numerous small details.}
%     \label{fig:failures}
% \end{figure*}

\begin{figure*}
    \centering
    \includegraphics[width=\linewidth]{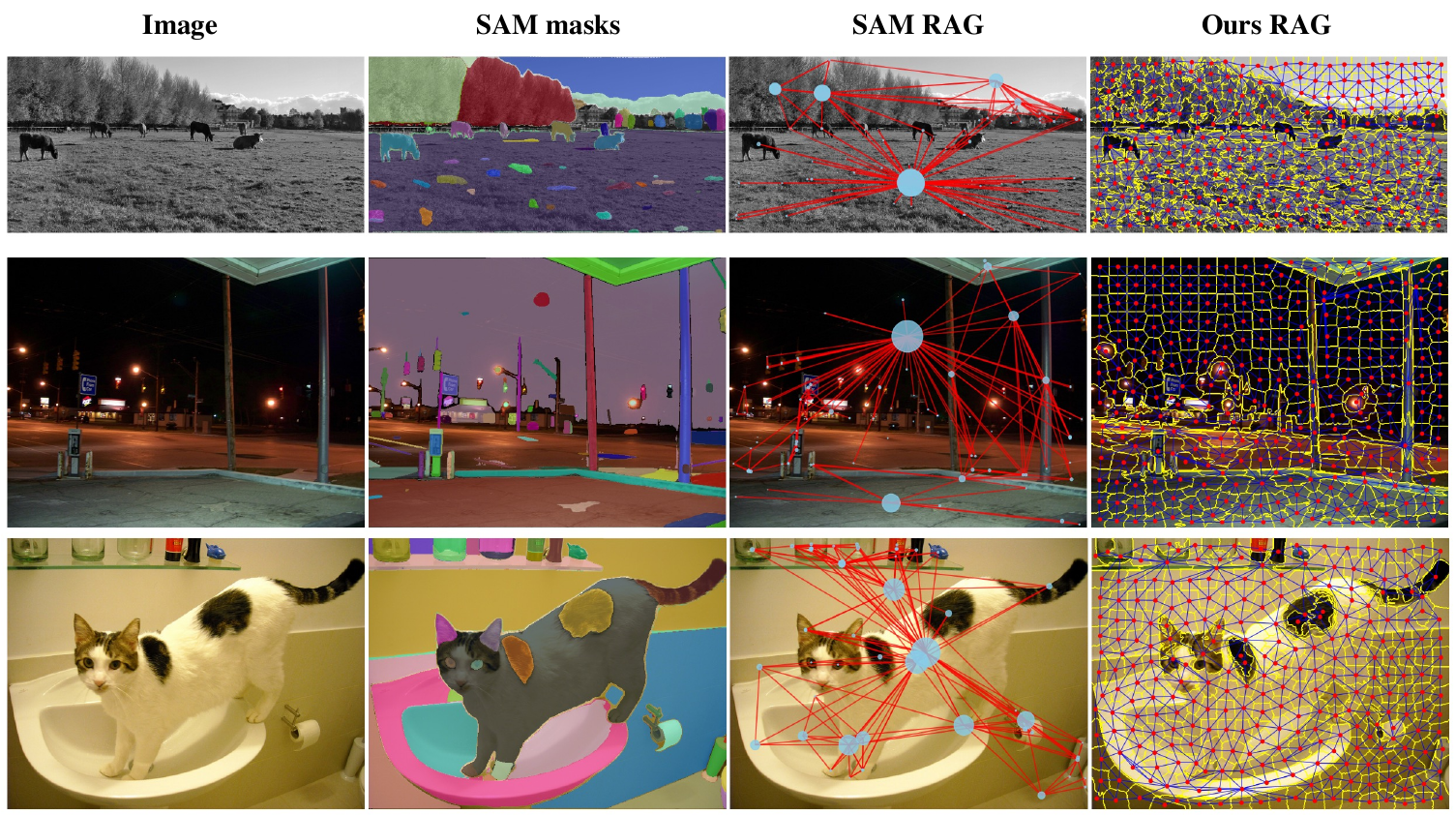}
    \caption{Comparison of RAG construction methods. While SAM masks (column 2) provide semantic regions, their imbalanced sizes create a problematic RAG (column 3). Large masks become hub nodes with numerous connections, dominating inter-region calculations; in column 3, larger circles indicate a greater number of neighbours. Our approach uses superpixels (column 4) to create a RAG with uniformly sized regions and a consistent number of neighbours, ensuring a more stable and reliable graph structure.}
    \label{fig:sam_rag}
\end{figure*}

\section{Combining SAM with our method}\label{s5}

The construction of the Region Adjacency Graph (RAG) is critical to our method's success. A natural consideration was to leverage powerful segmentation models like SAM to generate the graph's nodes. However, as Fig.~\ref{fig:sam_rag} reveals, this approach introduces a significant topological problem. SAM masks, while semantically meaningful, vary dramatically in size. This leads to the formation of a highly centralised RAG, where massive background regions become ``hub nodes'' that connect to a vast number of smaller regions (column 3). Such a structure is unstable for feature propagation, as these hubs can wash out important local details. To solve this, our method instead employs a superpixel-based tessellation of the image (column 4). This approach guarantees a granular and uniform partitioning, resulting in a balanced and regular RAG. Each node maintains local connectivity with a consistent number of neighbours, providing a stable and unbiased foundation for the structure-aware feature rectification.

% \begin{figure*}[!b]
%     \centering
%     \includegraphics[width=\linewidth]{src/challenge_vis.pdf}
%     \captionof{figure}{The effect of domain shift on segmentation performance.}
%     \label{fig:domain_shift}
% \end{figure*}

% \begin{figure*}[!b]
%     \centering
%     \includegraphics[width=\linewidth]{src/blur_vis.pdf}
%     \caption{The effect of texture destruction (gaussian blur) on segmentation performance.}
%     \label{fig:destruction}
% \end{figure*}

% \twocolumn[{%
% \renewcommand\twocolumn[1][]{#1}%
% \maketitle
% \begin{center}
%     \includegraphics[width=\linewidth]{src/blur_vis.pdf}
%     \captionof{figure}{The effect of texture destruction (gaussian blur) on segmentation performance.}
%     \label{fig:destruction}
% \end{center}%
% }]

% \begin{figure}
%     \centering
%     \includegraphics[width=\linewidth]{src/challenge_vis.pdf}
%     \caption{The effect of domain shift on segmentation performance.}
%     \label{fig:domain_shift}
% \end{figure}

% \input{sec/2_formatting}
% \input{sec/3_finalcopy}

{
    \small
    \bibliographystyle{ieeenat_fullname}
    \bibliography{main}

@inproceedings{clip,
  title={Learning transferable visual models from natural language supervision},
  author={Radford, Alec and Kim, Jong Wook and Hallacy, Chris and Ramesh, Aditya and Goh, Gabriel and Agarwal, Sandhini and Sastry, Girish and Askell, Amanda and Mishkin, Pamela and Clark, Jack and others},
  booktitle={International Conference on Machine Learning},
  pages={8748--8763},
  year={2021},
  organization={PMLR}
}

@inproceedings{maskclip,
  title={Extract free dense labels from clip},
  author={Zhou, Chong and Loy, Chen Change and Dai, Bo},
  booktitle={European Conference on Computer Vision},
  pages={696--712},
  year={2022},
  organization={Springer}
}

@inproceedings{tian2022generalized,
  title={Generalized few-shot semantic segmentation},
  author={Tian, Zhuotao and Lai, Xin and Jiang, Li and Liu, Shu and Shu, Michelle and Zhao, Hengshuang and Jia, Jiaya},
  booktitle={Proceedings of the IEEE/CVF Conference on Computer Vision and Pattern Recognition},
  pages={11563--11572},
  year={2022}
}

@article{reco,
  title={Reco: Retrieve and co-segment for zero-shot transfer},
  author={Shin, Gyungin and Xie, Weidi and Albanie, Samuel},
  journal={Advances in Neural Information Processing Systems},
  year={2022}
}

@article{ovdiff,
  title={Diffusion Models for Zero-Shot Open-Vocabulary Segmentation},
  author={Karazija, Laurynas and Laina, Iro and Vedaldi, Andrea and Rupprecht, Christian},
  journal={arXiv preprint arXiv:2306.09316},
  year={2023}
}

@article{voc12,
title={The pascal visual object classes challenge: A retrospective},
author={Everingham, Mark and Eslami, SM Ali and Van Gool, Luc and Williams, Christopher KI and Winn, John and Zisserman, Andrew},
journal={International Journal of Computer Vision},
year={2015}
}

@article{bai2024self,
  title={Self-calibrated clip for training-free open-vocabulary segmentation},
  author={Bai, Sule and Liu, Yong and Han, Yifei and Zhang, Haoji and Tang, Yansong},
  journal={arXiv preprint arXiv:2411.15869},
  year={2024}
}

@inproceedings{cliptrace,
  title={Explore the potential of clip for training-free open vocabulary semantic segmentation},
  author={Shao, Tong and Tian, Zhuotao and Zhao, Hang and Su, Jingyong},
  booktitle={European Conference on Computer Vision},
  pages={139--156},
  year={2024},
  organization={Springer}
}

@inproceedings{pascalcontext,
  title={The role of context for object detection and semantic segmentation in the wild},
  author={Mottaghi, Roozbeh and Chen, Xianjie and Liu, Xiaobai and Cho, Nam-Gyu and Lee, Seong-Whan and Fidler, Sanja and Urtasun, Raquel and Yuille, Alan},
  booktitle={Proceedings of the IEEE/CVF Conference on Computer Vision and Pattern Recognition},
  year={2014}
}

@inproceedings{naclip,
  title={Pay attention to your neighbours: Training-free open-vocabulary semantic segmentation},
  author={Hajimiri, Sina and Ayed, Ismail Ben and Dolz, Jose},
  booktitle={2025 IEEE/CVF Winter Conference on Applications of Computer Vision (WACV)},
  pages={5061--5071},
  year={2025},
  organization={IEEE}
}

@inproceedings{actualcoco,
  title={Microsoft coco: Common objects in context},
  author={Lin, Tsung-Yi and Maire, Michael and Belongie, Serge and Hays, James and Perona, Pietro and Ramanan, Deva and Doll{\'a}r, Piotr and Zitnick, C Lawrence},
  booktitle={European Conference on Computer Vision},
  pages={740--755},
  year={2014},
  organization={Springer}
}

@inproceedings{coco,
  title={Coco-stuff: Thing and stuff classes in context},
  author={Caesar, Holger and Uijlings, Jasper and Ferrari, Vittorio},
  booktitle={Proceedings of the IEEE/CVF Conference on Computer Vision and Pattern Recognition},
  year={2018}
}

@article{tian2020prior,
  title={Prior guided feature enrichment network for few-shot segmentation},
  author={Tian, Zhuotao and Zhao, Hengshuang and Shu, Michelle and Yang, Zhicheng and Li, Ruiyu and Jia, Jiaya},
  journal={IEEE transactions on pattern analysis and machine intelligence},
  volume={44},
  number={2},
  pages={1050--1065},
  year={2020},
  publisher={IEEE}
}

@inproceedings{peng2023hierarchical,
  title={Hierarchical dense correlation distillation for few-shot segmentation},
  author={Peng, Bohao and Tian, Zhuotao and Wu, Xiaoyang and Wang, Chengyao and Liu, Shu and Su, Jingyong and Jia, Jiaya},
  booktitle={Proceedings of the IEEE/CVF conference on computer vision and pattern recognition},
  pages={23641--23651},
  year={2023}
}

@inproceedings{huang2025revisit,
  title={Revisit the open nature of open vocabulary segmentation},
  author={Huang, Qiming and Hu, Han and Jiao, Jianbo},
  booktitle={Thirteenth International Conference on Learning Representations},
  year={2025}
}

@inproceedings{zegformer,
  title={Decoupling zero-shot semantic segmentation},
  author={Ding, Jian and Xue, Nan and Xia, Gui-Song and Dai, Dengxin},
  booktitle={Proceedings of the IEEE/CVF Conference on Computer Vision and Pattern Recognition},
  pages={11583--11592},
  year={2022}
}

@article{coop,
  title={Learning to prompt for vision-language models},
  author={Zhou, Kaiyang and Yang, Jingkang and Loy, Chen Change and Liu, Ziwei},
  journal={International Journal of Computer Vision},
  volume={130},
  number={9},
  pages={2337--2348},
  year={2022},
  publisher={Springer}
}

@inproceedings{cocoop,
  title={Conditional prompt learning for vision-language models},
  author={Zhou, Kaiyang and Yang, Jingkang and Loy, Chen Change and Liu, Ziwei},
  booktitle={Proceedings of the IEEE/CVF Conference on Computer Vision and Pattern Recognition},
  pages={16816--16825},
  year={2022}
}

@inproceedings{maple,
  title={Maple: Multi-modal prompt learning},
  author={Khattak, Muhammad Uzair and Rasheed, Hanoona and Maaz, Muhammad and Khan, Salman and Khan, Fahad Shahbaz},
  booktitle={Proceedings of the IEEE/CVF Conference on Computer Vision and Pattern Recognition},
  pages={19113--19122},
  year={2023}
}

@article{li2025closer,
  title={A closer look at the explainability of Contrastive language-image pre-training},
  author={Li, Yi and Wang, Hualiang and Duan, Yiqun and Zhang, Jiheng and Li, Xiaomeng},
  journal={Pattern Recognition},
  volume={162},
  pages={111409},
  year={2025},
  publisher={Elsevier}
}

@inproceedings{rpo,
  title={Read-only prompt optimization for vision-language few-shot learning},
  author={Lee, Dongjun and Song, Seokwon and Suh, Jihee and Choi, Joonmyeong and Lee, Sanghyeok and Kim, Hyunwoo J},
  booktitle={Proceedings of the IEEE/CVF International Conference on Computer Vision},
  pages={1401--1411},
  year={2023}
}

@article{clipood,
  title={CLIPood: Generalizing CLIP to Out-of-Distributions},
  author={Shu, Yang and Guo, Xingzhuo and Wu, Jialong and Wang, Ximei and Wang, Jianmin and Long, Mingsheng},
  journal={arXiv preprint arXiv:2302.00864},
  year={2023}
}

@article{clipsurgery,
  title={Clip surgery for better explainability with enhancement in open-vocabulary tasks},
  author={Li, Yi and Wang, Hualiang and Duan, Yiqun and Li, Xiaomeng},
  journal={arXiv preprint arXiv:2304.05653},
  year={2023}
}

@inproceedings{denseclip,
  title={Denseclip: Language-guided dense prediction with context-aware prompting},
  author={Rao, Yongming and Zhao, Wenliang and Chen, Guangyi and Tang, Yansong and Zhu, Zheng and Huang, Guan and Zhou, Jie and Lu, Jiwen},
  booktitle={Proceedings of the IEEE/CVF Conference on Computer Vision and Pattern Recognition},
  pages={18082--18091},
  year={2022}
}

@article{vit,
  title={An image is worth 16x16 words: Transformers for image recognition at scale},
  author={Dosovitskiy, Alexey and Beyer, Lucas and Kolesnikov, Alexander and Weissenborn, Dirk and Zhai, Xiaohua and Unterthiner, Thomas and Dehghani, Mostafa and Minderer, Matthias and Heigold, Georg and Gelly, Sylvain and others},
  journal={arXiv preprint arXiv:2010.11929},
  year={2020}
}

@article{sclip,
  title={SCLIP: Rethinking Self-Attention for Dense Vision-Language Inference},
  author={Wang, Feng and Mei, Jieru and Yuille, Alan},
  journal={arXiv preprint arXiv:2312.01597},
  year={2023}
}

@article{maskformer,
  title={Per-pixel classification is not all you need for semantic segmentation},
  author={Cheng, Bowen and Schwing, Alex and Kirillov, Alexander},
  journal={Advances in Neural Information Processing Systems},
  volume={34},
  pages={17864--17875},
  year={2021}
}

@inproceedings{mask2former,
  title={Masked-attention mask transformer for universal image segmentation},
  author={Cheng, Bowen and Misra, Ishan and Schwing, Alexander G and Kirillov, Alexander and Girdhar, Rohit},
  booktitle={Proceedings of the IEEE/CVF conference on computer vision and pattern recognition},
  pages={1290--1299},
  year={2022}
}

@inproceedings{groupvit,
  title={Groupvit: Semantic segmentation emerges from text supervision},
  author={Xu, Jiarui and De Mello, Shalini and Liu, Sifei and Byeon, Wonmin and Breuel, Thomas and Kautz, Jan and Wang, Xiaolong},
  booktitle={Proceedings of the IEEE/CVF Conference on Computer Vision and Pattern Recognition},
  pages={18134--18144},
  year={2022}
}

@article{catseg,
  title={CAT-Seg: Cost Aggregation for Open-Vocabulary Semantic Segmentation},
  author={Cho, Seokju and Shin, Heeseong and Hong, Sunghwan and An, Seungjun and Lee, Seungjun and Arnab, Anurag and Seo, Paul Hongsuck and Kim, Seungryong},
  journal={arXiv preprint arXiv:2303.11797},
  year={2023}
}

@inproceedings{clipes,
  title={Clip is also an efficient segmenter: A text-driven approach for weakly supervised semantic segmentation},
  author={Lin, Yuqi and Chen, Minghao and Wang, Wenxiao and Wu, Boxi and Li, Ke and Lin, Binbin and Liu, Haifeng and He, Xiaofei},
  booktitle={Proceedings of the IEEE/CVF Conference on Computer Vision and Pattern Recognition},
  pages={15305--15314},
  year={2023}
}

@article{clsclip,
  title={[CLS] Token is All You Need for Zero-Shot Semantic Segmentation},
  author={Wu, Letian and Zhang, Wenyao and Jiang, Tengping and Yang, Wankou and Jin, Xin and Zeng, Wenjun},
  journal={arXiv preprint arXiv:2304.06212},
  year={2023}
}

@inproceedings{mvpseg,
author = {Guo, Jie and Wang, Qimeng and Gao, Yan and Jiang, Xiaolong and Lin, Shaohui and Zhang, Baochang},
title = {MVP-SEG: Multi-view Prompt Learning for Open-Vocabulary Semantic Segmentation},
year = {2023},
publisher = {Springer-Verlag},
booktitle = {Pattern Recognition and Computer Vision: 6th Chinese Conference},
pages = {158–171},
}

@inproceedings{cityscapes,
  title={The cityscapes dataset for semantic urban scene understanding},
  author={Cordts, Marius and Omran, Mohamed and Ramos, Sebastian and Rehfeld, Timo and Enzweiler, Markus and Benenson, Rodrigo and Franke, Uwe and Roth, Stefan and Schiele, Bernt},
  booktitle={Proceedings of the IEEE conference on computer vision and pattern recognition},
  pages={3213--3223},
  year={2016}
}

@article{ade,
  title={Semantic understanding of scenes through the ade20k dataset},
  author={Zhou, Bolei and Zhao, Hang and Puig, Xavier and Xiao, Tete and Fidler, Sanja and Barriuso, Adela and Torralba, Antonio},
  journal={International Journal of Computer Vision},
  volume={127},
  pages={302--321},
  year={2019},
  publisher={Springer}
}

@inproceedings{cgformer,
  title={Contrastive grouping with transformer for referring image segmentation},
  author={Tang, Jiajin and Zheng, Ge and Shi, Cheng and Yang, Sibei},
  booktitle={Proceedings of the IEEE/CVF Conference on Computer Vision and Pattern Recognition},
  pages={23570--23580},
  year={2023}
}

@article{icar,
  title={icar: Bridging image classification and image-text alignment for visual recognition},
  author={Wei, Yixuan and Cao, Yue and Zhang, Zheng and Yao, Zhuliang and Xie, Zhenda and Hu, Han and Guo, Baining},
  journal={arXiv preprint arXiv:2204.10760},
  year={2022}
}

@inproceedings{openseed,
  title={A simple framework for open-vocabulary segmentation and detection},
  author={Zhang, Hao and Li, Feng and Zou, Xueyan and Liu, Shilong and Li, Chunyuan and Yang, Jianwei and Zhang, Lei},
  booktitle={Proceedings of the IEEE/CVF International Conference on Computer Vision},
  pages={1020--1031},
  year={2023}
}

@inproceedings{regionclip,
  title={Regionclip: Region-based language-image pretraining},
  author={Zhong, Yiwu and Yang, Jianwei and Zhang, Pengchuan and Li, Chunyuan and Codella, Noel and Li, Liunian Harold and Zhou, Luowei and Dai, Xiyang and Yuan, Lu and Li, Yin and others},
  booktitle={Proceedings of the IEEE/CVF Conference on Computer Vision and Pattern Recognition},
  pages={16793--16803},
  year={2022}
}

@inproceedings{proposalclip,
  title={Proposalclip: Unsupervised open-category object proposal generation via exploiting clip cues},
  author={Shi, Hengcan and Hayat, Munawar and Wu, Yicheng and Cai, Jianfei},
  booktitle={Proceedings of the IEEE/CVF Conference on Computer Vision and Pattern Recognition},
  pages={9611--9620},
  year={2022}
}

@inproceedings{dino,
  title={Emerging properties in self-supervised vision transformers},
  author={Caron, Mathilde and Touvron, Hugo and Misra, Ishan and J{\'e}gou, Herv{\'e} and Mairal, Julien and Bojanowski, Piotr and Joulin, Armand},
  booktitle={Proceedings of the IEEE/CVF international conference on computer vision},
  pages={9650--9660},
  year={2021}
}

@article{shi2024harnessing,
  title={Harnessing Vision Foundation Models for High-Performance, Training-Free Open Vocabulary Segmentation},
  author={Shi, Yuheng and Dong, Minjing and Xu, Chang},
  journal={arXiv preprint arXiv:2411.09219},
  year={2024}
}

@article{gui2024knn,
  title={kNN-CLIP: Retrieval enables training-free segmentation on continually expanding large vocabularies},
  author={Gui, Zhongrui and Sun, Shuyang and Li, Runjia and Yuan, Jianhao and An, Zhaochong and Roth, Karsten and Prabhu, Ameya and Torr, Philip},
  journal={arXiv preprint arXiv:2404.09447},
  year={2024}
}

@inproceedings{lan2024proxyclip,
  title={Proxyclip: Proxy attention improves clip for open-vocabulary segmentation},
  author={Lan, Mengcheng and Chen, Chaofeng and Ke, Yiping and Wang, Xinjiang and Feng, Litong and Zhang, Wayne},
  booktitle={European Conference on Computer Vision},
  pages={70--88},
  year={2024},
  organization={Springer}
}

@article{achanta2012slic,
  title={SLIC superpixels compared to state-of-the-art superpixel methods},
  author={Achanta, Radhakrishna and Shaji, Appu and Smith, Kevin and Lucchi, Aurelien and Fua, Pascal and S{\"u}sstrunk, Sabine},
  journal={IEEE transactions on pattern analysis and machine intelligence},
  volume={34},
  number={11},
  pages={2274--2282},
  year={2012},
  publisher={IEEE}
}

@inproceedings{kirillov2023segment,
  title={Segment anything},
  author={Kirillov, Alexander and Mintun, Eric and Ravi, Nikhila and Mao, Hanzi and Rolland, Chloe and Gustafson, Laura and Xiao, Tete and Whitehead, Spencer and Berg, Alexander C and Lo, Wan-Yen and others},
  booktitle={Proceedings of the IEEE/CVF international conference on computer vision},
  pages={4015--4026},
  year={2023}
}
}

\end{document}